%% file: main.tex
\documentclass[journal]{IEEEtran}

\usepackage{graphicx}
\usepackage{subfigure}
\usepackage{lipsum}
\usepackage{cite}
\usepackage{bm}
\usepackage{esvect}
\usepackage{amsmath}
\usepackage[ruled,vlined]{algorithm2e}
\usepackage{tabularx}
\usepackage{amsfonts}
\usepackage[normalem]{ulem}
\usepackage{makecell}
\usepackage{hyperref}

\graphicspath{
    {./fig/}
}

\usepackage{color}

\ifCLASSINFOpdf
\else
\fi

\newcommand{\revision}[2]{{#2}}

\begin{document}

\title{Planning Jerk-Optimized Trajectory with Discrete Time Constraints for Redundant Robots}

\author{Chengkai Dai$^{1,4}$, Sylvain Lefebvre$^{2}$, Kai-Ming Yu$^{3}$, Jo M.P. Geraedts$^{1}$ and Charlie C.L. Wang$^{4\dagger}$
\thanks{$^{1}$C. Dai and J.M.P. Geraedts are with the Department of Design Engineering, Delft University of Technology (TU Delft), Netherlands.} 
\thanks{$^{2}$S. Lefebvre is with INRIA, France.}
\thanks{$^{3}$K.-M. Yu is with the Department of Industrial and System Engineering, The Hong Kong Polytechnic University.}
\thanks{$^{4}$C. Dai and C.C.L. Wang are with the Department of Mechanical and Automation Engineering, The Chinese University of Hong Kong (CUHK).}
\thanks{This work was partially supported by the seed fund of Industrial Design Engineering faculty at TU Delft, the Natural Science Foundation of China (61628211), the CUHK Direct Grant (4055094) and a grant from the Research Grants Council of the Hong Kong SAR, China (Project No.: CUHK/14202219).}
\thanks{$^\dagger$Corresponding Author: {\tt\small cwang@mae.cuhk.edu.hk}}
}

\markboth{IEEE Transactions on Automation Science and Engineering,~Accepted}%
{Dai \MakeLowercase{\textit{et al.}}: Planning Jerk-Optimized Trajectory with Discrete Time Constraints for Redundant Robots}

\maketitle

\begin{abstract}
We present a method for effectively planning the motion trajectory of robots in manufacturing tasks, the tool-paths of which are usually complex and have a large number of discrete time constraints as waypoints. Kinematic redundancy also exists in these robotic systems. The jerk of motion is optimized in our trajectory planning method at the meanwhile of fabrication process to improve the quality of fabrication. 
Our method is based on a sampling strategy and consists of two major parts. After determining an initial path by graph-search, a greedy algorithm is adopted to optimize a path by locally applying adaptive filers in the regions with large jerks. The \revision{filtering}{filtered} result is obtained by numerical optimization. In order to achieve efficient computation, an adaptive sampling method is developed for learning a collision-indication function that is represented as a support-vector machine. Applications in robot-assisted 3D printing are given in this paper to demonstrate the functionality of our approach.
\end{abstract}

\textit{Note to Practitioner}
\begin{abstract}
In robot-assisted manufacturing applications, robotic arms are employed to realize the motion of workpieces (or machining tools) specified as a sequence of waypoints with the positions of tool tip and the tool orientations constrained. The required \revision{degree-of-freedoms (DOFs) are}{degree-of-freedom (DOF) is} often less than the robotic hardware system (e.g., a robotic arm has \revision{6-DOFs}{6-DOF}). Specifically, rotations of the workpiece around the axis of a tool can be arbitrary (see Fig.~\ref{figPath} for an example). By using this redundancy –- i.e., there are many possible poses of a robotic arm to realize a given waypoint, the trajectory of robots can be optimized to consider the performance of motion in velocity, acceleration and jerk in the joint space. In addition, when fabricating complex models each tool-path can have a large amount of waypoints. It is crucial for a motion planning algorithm to compute a smooth and collision-free trajectory of robot to improve fabrication quality. 
The time taken by the planning algorithm should not significantly lengthen the total manufacturing time; ideally it would remain hidden as computing motions for a layer can be done while the previous layer is printing. The method presented in this paper provides an efficient framework to tackle this problem. 
The framework has been well tested on our robot-assisted additive manufacturing system to demonstrate its effectiveness and can be generally applied to other robot-assisted manufacturing systems.
%
\end{abstract}

\begin{figure}[t!]
\centering
\includegraphics[width=\linewidth]{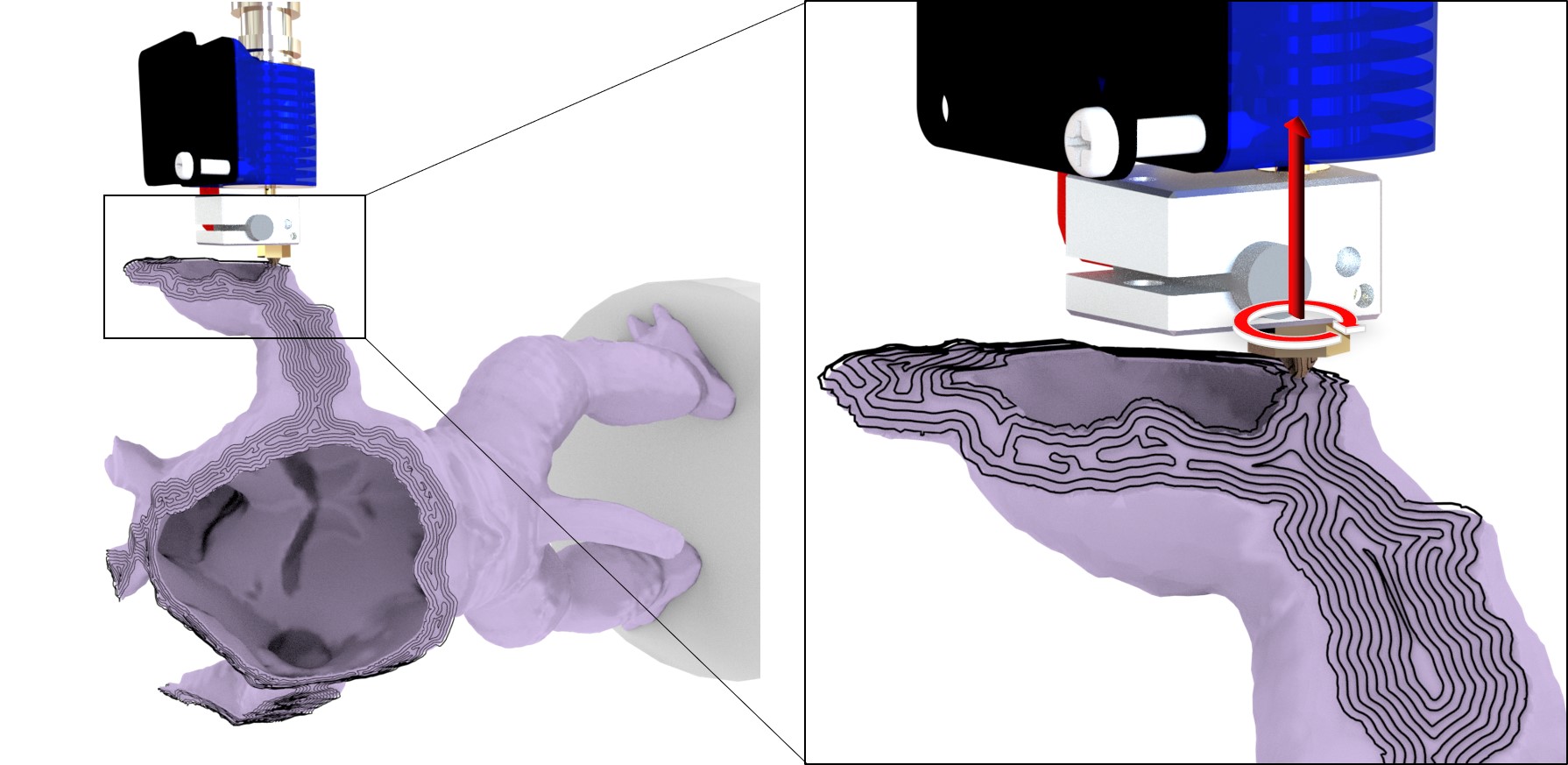}
\caption{An example tool-path for robot-assisted 3D printing system \cite{dai18}, rotation around the red axis can be freely changed because of kinematic redundancy.}
\label{figPath}
\end{figure}


\input{secIntroductionNew}

\input{secRelatedWorks}

\input{secTrajectoryPlanning}

\input{secCollisionClassifierNew}

\input{secExperimentResult}

\section{Conclusion and Discussion}\label{secConclusion}
We have presented a novel sampling-based framework for planning discrete time constrained trajectories on redundant robots. The major technical contributions include 1) a local filter for jerk minimization while considering other hardware-oriented constraints for feasibility, 2) a greedy algorithm to be applied to a path with many waypoints and 3) an adaptive sampling strategy for effectively learning a collision-indication function with high accuracy. With the help of these techniques, our approach can efficiently minimize the total jerk and reduce the maximal jerk. 
\revision{The}{Our} motion planning solution \revision{that we can achieve}{} is $40 \times$ faster \revision{}{than the method in \cite{dai18} when being applied to all tool-paths of the Armadillo model -- the model shown in Fig.\ref{figPath}}.

We have tested the performance of jerk-minimized trajectory in the application of robot-assisted 3D printing using a setup with one robotic arm and one tilting table. The results of our experiment tests are very encouraging, where the fabrication quality in terms of smoothness \revision{is}{has been} clearly improved while the time efficiency is ensured. 
\revision{}{However, the experiments are still conducted on hardware systems with relative low DOF (i.e., $L \leq 8$). The scalability of our algorithm on higher DOF needs to be further investigated. Another limitation of our method is that the collision indication function obtained by learning assumes a very certain environment. It will be interesting to see how it can be further developed for a dynamic environment -- e.g., human-robot interaction.}  
For future development, we will investigate the relationship between motion smoothness and layer-height / path-width in additive manufacturing. 
Moreover, we also plan to test this approach of jerk-optimized trajectory planning in the applications of machining (i.e., subtractive manufacturing).

\section*{Acknowledgement}
Authors would thank the help offered by Tianyu Zhang, Xiangjia Chen and Allan Mok during the physical experiments of this research.

\ifCLASSOPTIONcaptionsoff
  \newpage
\fi

\bibliographystyle{IEEEtran}
\bibliography{references}

\enlargethispage{-6.5cm}

\begin{IEEEbiography}[{\includegraphics[width=1in,height=1.25in,clip,keepaspectratio]{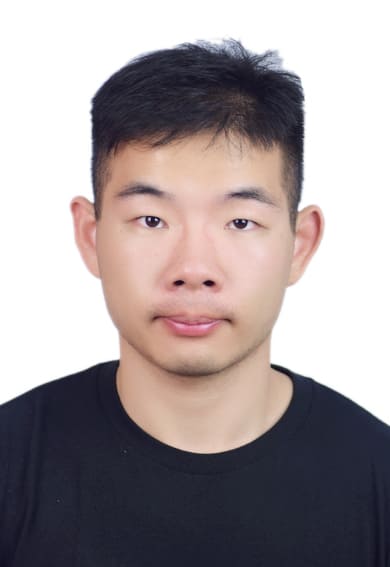}}]{Chengkai Dai}
is currently a Ph.D. candidate of the Department of Sustainable Design Engineering at Delft University of Technology. His research area includes robotics, geometry computing and computational design.
\end{IEEEbiography}

\begin{IEEEbiography}[{\includegraphics[width=1in,height=1.25in,clip,keepaspectratio]{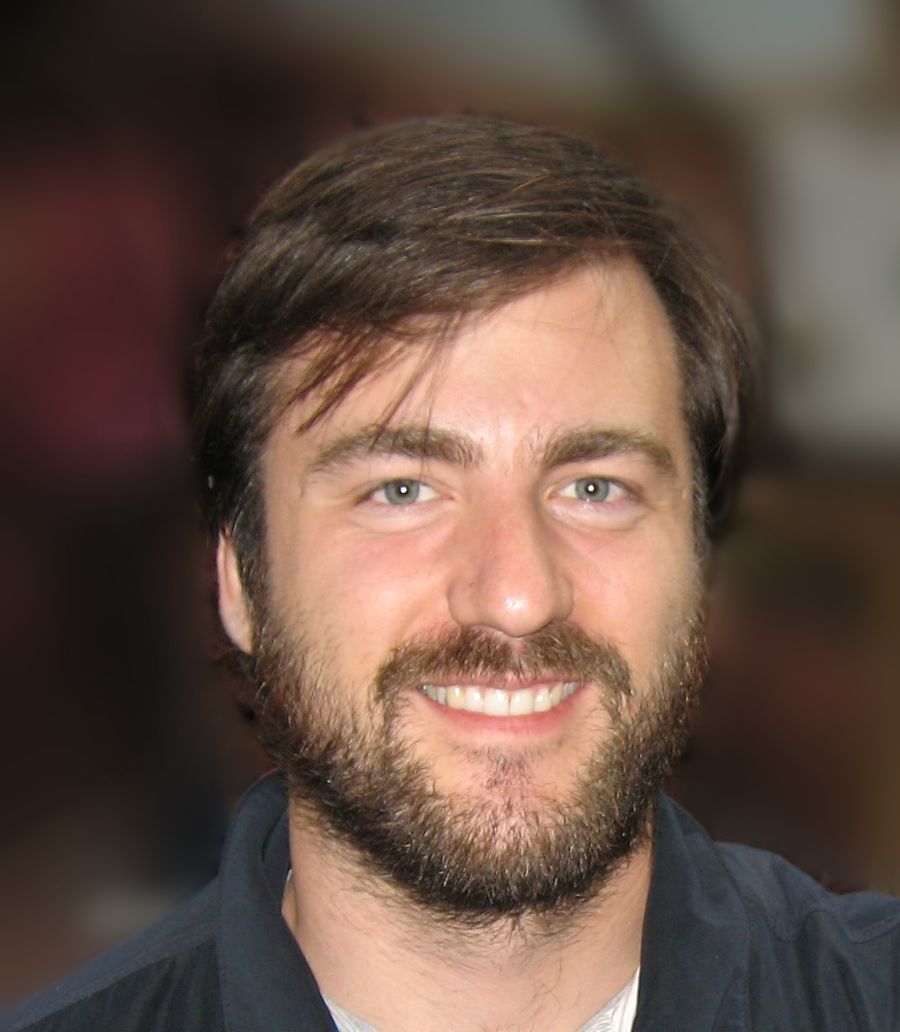}}]{Sylvain Lefebvre}
  is a senior researcher at Inria (France), where he leads the MFX team. His main research focus is to simplify content creation by automatically synthesizing highly detailed patterns, structures and shapes, with applications in Computer Graphics and Additive Manufacturing. Sylvain received the EUROGRAPHICS Young Researcher Award in 2010. From 2012 to 2017 he was the principal investigator of the ERC ShapeForge (StG) and IceXL (PoC) projects. He created and is the lead developer of the IceSL software for modeling for additive manufacturing.
\end{IEEEbiography}

\begin{IEEEbiography}[{\includegraphics[width=1in,height=1.25in,clip,keepaspectratio]{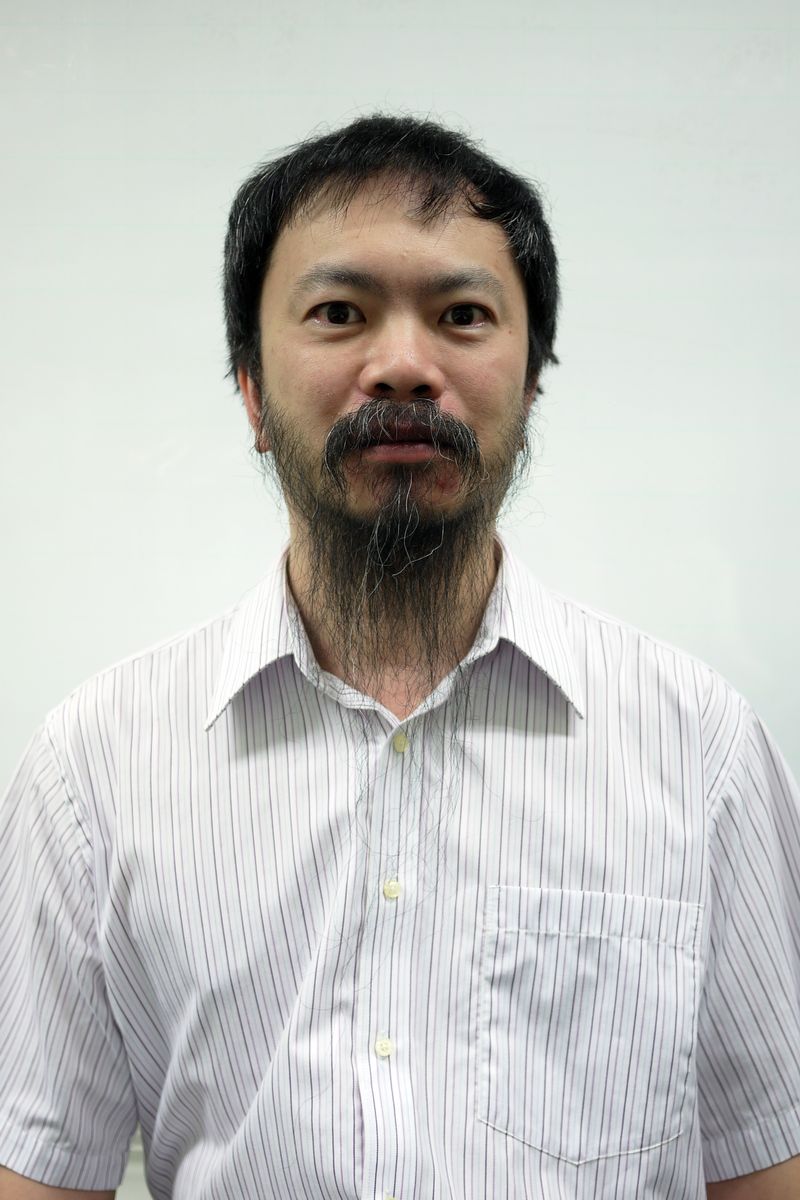}}]{Kai-Ming Yu}
  received his BSc (Eng) in Mechanical Engineering from the University of Hong Kong in 1985. He obtained his PhD from the University of Hong Kong, Department of Mechanical Engineering in 1991. He worked in the Research Centre and Mechanical Engineering Department of the Hong Kong University of Science and Technology until 1993. He is currently an Assistant Professor in the Hong Kong Polytechnic University Industrial and Systems Engineering Department. His research interests include CAD/CAM, CAE, and PDM, reverse engineering and rapid prototyping technologies. He is also a Senior Member of the Society of Manufacturing Engineers.
\end{IEEEbiography}

\begin{IEEEbiography}[{\includegraphics[width=1in,height=1.25in,clip,keepaspectratio]{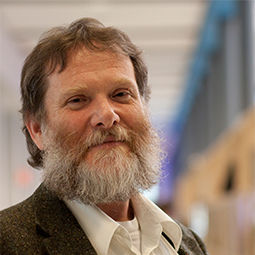}}]{Jo M.P. Geraedts}
  received the M.Sc. degree from Eindhoven University of Technology, Eindhoven, Netherlands, in 1976, and the Ph.D. degree from University of Nijmegen, Nijmegen, Netherlands, in 1983, both in physics. From 1983 to 2017, he was with Oc\'{e}-van der Grinten N.V. Dr. Geraedts is the Emeritus Professor of New Mechatronic Design at TU Delft. His research areas include 3-D scanning, 3-D printing, and human–robot interaction.
\end{IEEEbiography}

\begin{IEEEbiography}[{\includegraphics[width=1in,height=1.25in,clip,keepaspectratio]{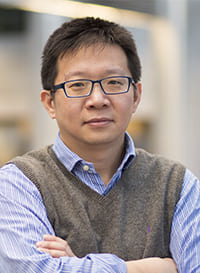}}]{Charlie C.L. Wang}
is currently a Professor of Mechanical and Automation Engineering and Director of Intelligent Design and Manufacturing Institute at the Chinese University of Hong Kong (CUHK). Before that, he was a tenured Professor and Chair of Advanced Manufacturing at Delft University of Technology (TU Delft), The Netherlands. He received the Ph.D. degree (2002) from Hong Kong University of Science and Technology in mechanical engineering, and is now a Fellow of the American Society of Mechanical Engineers (ASME) and the Hong Kong Institute of Engineers (HKIE). His research areas include geometric computing, intelligent design and advanced manufacturing.
\end{IEEEbiography}

\vfill
\end{document}

%% file: secIntroductionNew.tex
\section{Introduction}\label{secIntroduction}
\IEEEPARstart{R}{obotic} arms are increasingly used as production tools to realize customized manufacturing processes in the applications of Industry 4.0.
We focus on cases where a robotic arm is used to produce a physical part from a digital model, following a set of tool-paths generated by the process-planning algorithm.
%
The trajectory of motion is often optimized for a robotic manipulator with kinematic redundancy -- e.g., a six \textit{degree-of-freedom} (DOF) robotic arm is employed to realize a trajectory with 5-DOF constraints (see Fig.~\ref{figPath} for an example in additive manufacturing \cite{dai18}). Other than additive manufacturing, example processes include robotic machining \cite{hu99}, automated tape laying for composite fabrication \cite{marsh11}, etc. 

\revision{Trajectory planning in these applications}{For these applications, trajectory planning} needs to commonly satisfy the following requirements.
\begin{itemize}
\item \textit{Discrete time constraints}: The target trajectories are usually represented as a set of waypoints with given positions and orientations\footnote{Orientation is given as a unit vector so that only 5-DOFs are constrained.} to be accurately passed through at the tip of tool installed on the robot end-effector. Moreover, speed of the tool is also controlled by assigning a time parameter to each waypoint. Note that the speed for material removal (in subtractive manufacturing \cite{chan15}) or the speed for material accumulation (in additive manufacturing \cite{etienne19,Ibrahim16}), called \textit{feedrate}, is a very important parameter to be controlled in fabrication. 

\item \textit{Optimized jerk}: The task of tool-path with discrete time constraints is mapped into the joint space of a robotic manipulator by its inverse kinematics. As the time derivative of acceleration in joint space, jerk has great influence on the smoothness of a manipulator's motion. A motion with smaller jerk has less vibration. To reduce the vibration into a low level is crucial for realizing a high quality robotic fabrication. An ideal trajectory should have an integral-norm of the jerk minimized and the maximal jerk controlled below a reasonable bound.

\item \textit{Collision-free}: Collisions with the surrounding objects and the parts of a model that have already been manufactured must be avoided at all costs along the trajectory of the robot system. 
While optimizing the jerk, a collision-free property needs to be ensured along the optimized trajectory. This becomes a computational bottleneck when the obstacles have a complex shape. This is unfortunately the case for robotic fabrication as the models to be fabricated usually have complex 3D freeform surfaces.
\end{itemize}
%
Generating a trajectory satisfying the aforementioned requirements is challenging.
The tool-paths for robotic fabrication can contain a large number of waypoints (typically in the thousands), which are hard to handle with global methods \cite{TrajOpt13}. Online (local) planning methods (e.g., \cite{Martin89, Siciliano1990}) often include time-jerk optimization strategies to improve the quality of the trajectories. 
%
However, they are not applicable to discrete time constraints. 
A sampling-based framework is proposed in this paper to tackle this problem effectively and efficiently. 

\subsection{Problem Statement}
Suppose a robotic manipulator with $L$ DOFs ($L>5$) is employed to follow a user specified path $\mathbf{x}(t) \in \mathbb{R}^5$ with the tool tip held by its end-effector, the location of which is determined by parameters in joint space (i.e., $\mathbf{q}(t) \in \mathbb{R}^L$) by the forward kinematics as:
\begin{center}
$f(\mathbf{q}):\mathbf{q}\in \mathbb{R}^L \mapsto \mathbf{x}\in \mathbb{R}^5$.
\end{center}
Given a collision-indication function $\Gamma(\mathbf{q})$ the sign of which indicates a collision-occurring (`$+$') or collision-free (`$-$') configuration, we can define the collision-free configuration space as:
\begin{center}
$\mathcal{C}_{free}=\{ \mathbf{q} \; | \; \Gamma(\mathbf{q})<0, \forall \mathbf{q} \in \mathbb{R}^L\}$.
\end{center}
\revision{The problem of finding a feasible minimum-jerk trajectory following the task tool-paths is defined as follows}{}
\revision{where the measure of jerk is $\| \dddot{\mathbf{q}} \|^2_W = \mathbf{q}^T \mathbf{W} \mathbf{q}$ with $\mathbf{W}$ being a non-negative matrix giving the weights for relative importance between the joints.}{} In robotic fabrication, \revision{this results in}{the problem of finding a feasible jerk-minimized trajectory following the task tool-paths is defined as} a complex trajectory planning problem with \revision{large inputs}{a large size of input} (i.e., many waypoints along freeform surfaces as discrete time constraints). \revision{We therefore propose a sampling-based strategy.}{} 
Given a set of $M$ waypoints each at a prescribed time $t_i$ with the position of tool tip $\mathbf{p}_i$ and the tool orientation $\mathbf{n}_i$, denoted as
\begin{equation}\label{eqWaypoints}
\mathbf{x}(t_i) = (\mathbf{p}_i,\hat{\mathbf{n}}_i) \qquad (\forall i = 1,\ldots,M),
\end{equation}
we formulate the optimization problem to be solved as 
\begin{equation}\label{eqGlobalOptm}
\begin{aligned}
            & \arg \min_{\{ \mathbf{q}_i\} } \mathbb{J}=\sum_{i=1}^M \| \dddot{\mathbf{q}}(t_i) \|_W^2 \\
s.t. \quad  & \mathbf{x}(t_i)=f(\mathbf{q}(t_i)) \quad (\forall i=1,\ldots,M), \\
            & \Gamma(\mathbf{q}(t_i))<0, \\
            & \mathbf{q}_{\min} \leq \mathbf{q}(t_i) \leq \mathbf{q}_{\max}, \\ 
            & |\dot{\mathbf{q}}(t_i)| \leq \mathbf{v}_{\max}, \, |\ddot{\mathbf{q}}(t_i)| \leq \mathbf{a}_{\max}, \, |\dddot{\mathbf{q}}(t_i)| \leq \mathbf{j}_{\max}. 
\end{aligned}
\end{equation}
Here \revision{}{the measure of jerk is $\| \dddot{\mathbf{q}} \|^2_W = \dddot{\mathbf{q}}^T \mathbf{W} \dddot{\mathbf{q}}$ with $\mathbf{W}$ being a non-negative diagonal matrix giving the weights for relative importance between the joints which can also be assigned as equal importance by $\mathbf{W}=\mathbf{I}$, and} $| \cdot |$ returns a vector with the absolute value of every component. The last four constraints (i.e., the last two lines in Eq.(\ref{eqGlobalOptm})) are about joint's position, velocity, acceleration and jerk, and are defined according to the hardware limits. $\hat{\mathbf{n}}_i$~is a normalized vector so that only 5-DOFs are constrained. For the sake of compact notation, we denote $\mathbf{x}(t_i)$ and $\mathbf{q}(t_i)$ as $\mathbf{x}_i$ and $\mathbf{q}_i$ in the rest of this paper. \revision{Different weights can be given to the diagonal elements of $\mathbf{W}$ to specify the different importance for different joints, and it can also be simply assigned as equal importance by $\mathbf{W}=\mathbf{I}$.}{}

Note that the quality and feasibility of a trajectory is evaluated at discrete time samples in our formulation. We argue that a weak form solution (i.e., resolution completeness) for the \revision{minimum-jerk}{jerk-optimized} trajectory planning is obtained when the sampling points are dense enough. Similar strategies have been used and adopted by the robotics community for motion planning \cite{cheng02,Karaman2011SAO}. 

\subsection{Our Approach}
\revision{}{Directly solving the jerk optimization problem along a trajectory with discrete time constraints is time-consuming even if the state-of-the-art method such as \cite{TrajOpt13} is employed.} 
\revision{We}{In a relaxed formulation, we} propose a greedy algorithm based on \textit{local adaptive filtering} to the jerk \revision{along a trajectory with discrete-time constraints.}{after computing an initial trajectory that minimizes the total velocity variation.} 
We overcome the computational bottleneck of collision detection by a learning-based collision estimator that approximates the continuous decision function.

\begin{itemize}
\item An adaptive greedy algorithm to generate jerk-optimized trajectory with discrete-time constraints (Section \ref{secPlanning});

\item An adaptive sampling strategy for effectively learning a collision-indication function (Section \ref{secCollision}).
\end{itemize}
In summary, we develop a new sampling-based framework for planning discrete time constrained trajectory on redundant robots, which can effectively and efficiently generate jerk-optimized \revision{trajectory}{trajectories} for robotic fabrication.

%% file: secRelatedWorks.tex
\section{Related work}\label{secRelatedWorks}
In this section, we review the prior research related to the two major parts of our framework: the trajectory planning approaches considering the task-oriented constraints and the machine-learning based collision-detection approaches.

\subsection{Trajectory Planning for Task-Oriented Constrains}
Many robot-assisted manufacturing tasks impose constraints on the robot's motion that exhibit kinematic redundancy, where more DOFs are available than the needed DOFs to realize the task. An example is to accumulate materials in 3D printing by tracing a given path with an axi-symmetric filament extruder \cite{dai18,wu17}. In robotic motion planning, there are two types of redundancy, intrinsic and functional ones. 
Intrinsic redundancy occurs when the dimension of the joint space spanned by a robot's joint variables is greater than the dimension of its operational space, which is the reachable Cartesian space of the end-effector. 
Functional redundancy is the case where the dimension of the robot's operational space is greater than the dimension of the task space (e.g., the waypoints to be realized).

The intrinsic redundancy problem has been discussed extensively by robotic researchers. Most of the existing methods are playing with the null space of the Jacobian matrix, since the Jacobian matrix is non-square, and exploit the self-motion space of redundant robots\cite{Nakamura1990} \cite{Yoshikawa96}. Functional redundancy is different from intrinsic redundancy, in which the Jacobian matrix is non-singular and square. \revision{There is no chance to get}{It is not possible to obtain} the null space of the robot itself. Huo and Bason \cite{Huo08} proposed to add an extra column to the Jacobian matrix by introducing a virtual joint. After that, a general solution called \textit{twist decomposition algorithm} was introduced by using the projection matrices in the operational space to find the null space of Jacobian.

Existing generic trajectory planning methods approach functional redundancy by local \cite{Martin89, Siciliano1990} or global \cite{Seereeram93} optimization techniques. The optimization is usually based on different objective metrics such as avoiding obstacles~\cite{Khadem91}, avoiding joint limits~\cite{Huo08}, \revision{avoid}{avoiding} singularities \cite{LEGER2016155} and/or minimizing joint velocities~\cite{Dubey88}, jerks~\cite{Freeman12} and torques. Some \revision{approach \cite{Huo08} optimizes}{approaches (e.g., \cite{Huo08,Zhang19}) optimize} the combination of multiple criteria.
\begin{itemize}
\item In local optimization, the strategy is to generate joint configurations  \cite{Martin89, Siciliano1990} that minimize the instantaneous value of the aforementioned metrics. However, these optimization methods only guarantee that a local minima of the objective function is found, which may not be sufficient to ensure path continuity as a whole.

\item Global optimization seeks to generate trajectories that minimize the integral of the performance metric over a prescribed interval, as opposed to just instantaneously 
in time\revision{ (e.g., the TrajOpt approach~\cite{TrajOpt13})}{}. However, these approaches are time-consuming 
because \revision{of the exponential cost for computing optimal solutions}{that the geometric algorithm for obstacle avoidance has been included in the loop of computation (ref. \cite{TrajOpt13,Zhang19})}. 
The optimizations also suffer from the initial guess problem – i.e., whether solutions can be found highly relies on the initial guess. Thus, existing global approaches cannot guarantee an algorithmic completeness (some inputs may not lead to solutions).
\end{itemize}
\revision{Instead}{Differently}, our approach is sampling-based and can ensure algorithmic completeness.
%
%

In the area of motion planning, sampling-based algorithms are the most successful method because of their efficiency and completeness \cite{Kavraki96,Lavalle00}. To deal with the redundancy, researchers define constraints as manifolds and efficiently sample the manifolds by forcing the constraints via rejection sampling and projection sampling of the configuration space \cite{Stilman10,Berenson09}. However, such algorithms cannot handle the tracking problem with a predefined path. Our work is partially similar with a new global graph search method \cite{Qiang2018}, where the self-motion space is parameterized by angular and path distances and a graph is constructed by cell-decomposition applied to these two parameters. The trajectory planning problem is solved by a shortest path search on the graph. However, the method can be very slow when a dense sampling is applied. In addition, it is unclear how to optimize for minimal jerkiness by such a graph-search based method.

\subsection{Machine-Learning Based Collision Detection}
When performing the trajectory planning in the configuration space, the solutions falling in the sub-space of collision-occurring configurations should be abandoned. For realizing the collision avoidance, the collision-free configuration space should be computed and represented in an appropriate way. As discussed in \cite{Pan15}, geometry-based methods are usually limited to low-dimensional configuration spaces, due to the combinatorial complexity involved in computing the boundary of the collision-free space for high-dimensional configurations. Moreover, for computing the optimal trajectory in a numerical optimization framework, the collision-free configuration space is desired to be presented (or approximated) by an algebraic function (i.e., $\Gamma(\cdot)$ as discussed above in  Eq.(\ref{eqGlobalOptm})).

To solve this problem, machine learning techniques have been used for collision detection to approximate collision-free spaces based on sampled configurations. For example, Pan et al. \cite{Pan13} conducted the incremental \textit{support vector machines} (SVM) to learn a representation of configuration space in an online step. Their method samples the configuration space by iteratively exploiting the near boundary configurations. 
Das et al. \cite{Das17} developed a kernel-based perceptron learning algorithm which can efficiently update the classifier actively. This enables the function to online update the decision boundary of a classifier. 
\textit{Gaussian mixture models} (GMM) are used in \cite{Huh16} to represent the collision-aware configuration space, from which the collision detection is performed by assigning a query configuration with the same label as the closest Gaussian. 
Pan and Manocha \cite{doi:Pan16} adopt a \textit{k-nearest neighbor} (k-NN) model in their sampling-based motion planners, which can significantly reduce the time required for collision checking. 
Neural networks also have been applied to perform collision checking (e.g., \cite{Garcia02}); however, the training step could be time-consuming when multiple-layers need to be trained in a neural network. 
Recently, Salehian et al.~\cite{Seyed18} develop an exhaustive sampling method to find a collision decision function, which can be treated as collision constraints in the optimization-based computation of inverse kinematics. 
Although potentially applicable after certain modifications, none of the above approaches have considered the specific situation in robot-assisted manufacturing where the shape of obstacles (specifically the workpiece to work on) is changing with time. For instance in additive manufacturing the part being created is making the space of collision-free configurations increasingly complex. In this paper, we develop 
{a sophisticate} method to tackle this manufacturing-oriented situation, which can use a very small number of training samples to provide an accurate function for efficiently estimating the collision-indication function in the numerical optimization framework.

%% file: secTrajectoryPlanning.tex
\section{Trajectory Planning}\label{secPlanning}
Given a set of waypoints as discrete time constraints for robotic fabrication, our trajectory planning algorithm computes the configurations of manipulators as a sequence of optimized configurations $\mathcal{Q}=\{ \mathbf{q}_i\}$ ($i=1,\ldots,M$) in two steps. Firstly, a trajectory is determined by a graph-search method that minimizes the total cost of joint transition -- i.e., the initial values of $\mathbf{q}_i$s 
are assigned. In the second step, a greedy algorithm is developed to optimize the trajectory by locally applying adaptive filters to adjust the value of $\mathbf{q}_i$s 
in regions with large jerks. A validation mechanism is developed to ensure the resultant trajectory is completely collision-free at the waypoints. 

\subsection{Initialization}\label{subsecInit}
We employ a sampling-based method to determine a feasible trajectory. 
When the sampling rate is dense enough, it provides a very good solution for \revision{determine}{determining} an initial trajectory to be further optimized.

Since only position and orientation are defined on a waypoint, the tool is allowed to rotate freely around the tool axis orientation $\hat{\mathbf{n}}_i$. Therefore, this leads to infinite possibilities to define a pose in the robotic manipulator's configuration space, which results in kinematic redundancy. The rotation can be defined by a quaternion as a rotational angle $\theta$ around the vector $\hat{\mathbf{n}}_i$, that is
\begin{equation}\label{eqQuaternion}
\mathbf{h}_i =(\hat{\mathbf{n}}_i \sin(\frac{\theta}{2}), \cos(\frac{\theta}{2})) \quad (\forall \theta \in [-\pi, \pi]).
\end{equation}
We first sample the task space at every waypoint by using different values of $\theta$, and then employ a graph-search approach to find a feasible path by connecting the selected samples -- one from each waypoint. 

\begin{figure}[t!]
\centering
\includegraphics[width=\linewidth]{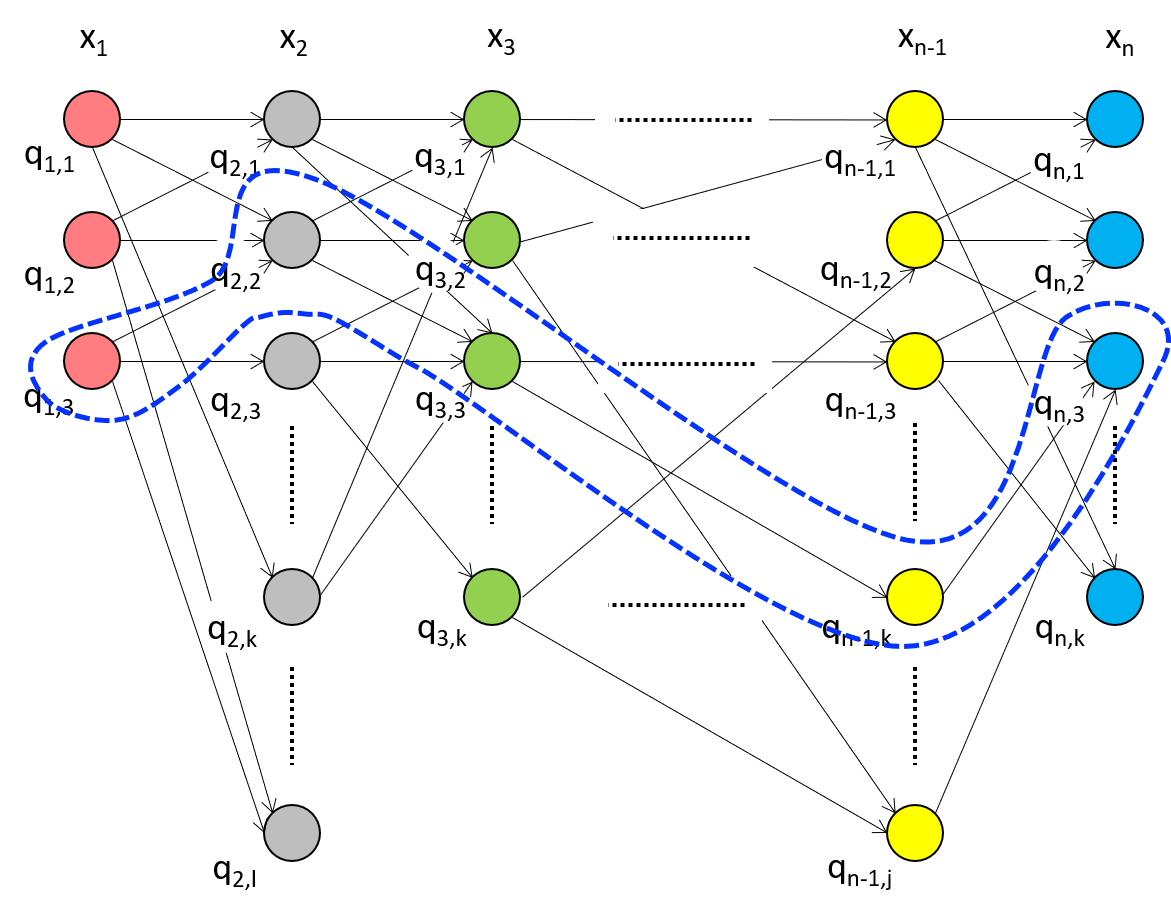}
\caption{An illustration for the graph used in our approach to find an initial trajectory. Nodes in the same column (called ladder) represent the different feasible solutions in the joint space for realizing the same way point. Edges are added between nodes in neighboring ladders. The shortest path on the graph is highlighted by \revision{bold lines and the red nodes are the corresponding solutions for every waypoints in the joint space.}{the blue dashed lines.}
}\label{figGraph}
\end{figure}

A graph $\mathcal{G}$ spanning the task space of a given path $\mathcal{X}=\{ \mathbf{x}_i\}$ is constructed by the following method (see also Fig.~\ref{figGraph} for an illustration). 
\begin{itemize}
\item 
\textit{Nodes}: Each waypoint $\mathbf{x}_i=(\mathbf{p}_i, \hat{\mathbf{n}}_i)$ is first uniformly sampled into $n$ rotational angles for $\theta \in [-\pi, \pi]$ to determine $n$ quaternions by Eq.(\ref{eqQuaternion}). This leads to $n$ points in the special Euclidean group $\mathbb{SE}(3)$. For each point in $\mathbb{SE}(3)$ multiple kinematic solutions in the joint space can be determined by \textit{inverse kinematics} (IK). Without loss of generality, we assume that $N_i$ points ($N_i>n$) in the joint space can be obtained for realizing a waypoint $\mathbf{x}_i$ -- denoted by $\mathbf{q}_{i,j}$ ($j=1,\ldots,N_i$). They are defined as a ladder of nodes, $\mathcal{G}_i=\{ \mathbf{q}_{i,j} \}$, in $\mathcal{G}$. Each ladder of nodes is displayed as a column of nodes in Fig.~\ref{figGraph}. Nodes corresponding to a configuration with collision are excluded from the graph, which can be efficiently checked by a collision detection library (e.g., \cite{Pan12}). 
\item \textit{Edges}: Directed edges are constructed by linking nodes in a ladder $\mathcal{G}_i$ to nodes in the next ladder $\mathcal{G}_{i+1}$ while respecting the joint velocity limits. The edge between $\mathbf{q}_{i,j}$ and $\mathbf{q}_{i+1,k}$ is only added when 
\begin{equation}
\frac{\mathbf{q}_{i+1,k}-\mathbf{q}_{i,j}}{t_{i+1}-t_i} \leq \mathbf{v}_{\max}.
\end{equation}
\revision{}{To avoid the `winding effect' caused by revolute joints, we evaluate the circular distance between configurations here and also the rest of this paper.} 
The following transition cost is added as the weight of an edge (angular-velocity estimation):
\begin{equation}
w(\mathbf{q}_{i,j},\mathbf{q}_{i+1,k})=(\mathbf{q}_{i+1,k} -  \mathbf{q}_{i,j})^T \mathbf{W} (\mathbf{q}_{i+1,k} -  \mathbf{q}_{i,j}) 
\end{equation}
with $\mathbf{W}$ being the non-negative \revision{}{diagonal} matrix that gives the weights of relative importance between the joints.
\end{itemize}
A shortest path on $\mathcal{G}$ from a start node $\mathbf{q}_s \in \mathcal{G}_{1}$ to an end node $\mathbf{q}_e \in \mathcal{G}_{M}$ actually defines a trajectory $\mathcal{Q}$ that minimizes the total cost of transition 
as follows
\begin{equation}\label{eqOptmTrans}
\mathbb{J}_{trans}=\sum_{i=1}^{M-1} \| \mathbf{q}_{i+1} - \mathbf{q}_{i} \|_W^2.
\end{equation}

The shortest path $\mathcal{P}$ on $\mathcal{G}$ can be found by applying the Dijkstra's algorithm for multiple sources -- using all nodes in $\mathcal{G}_1$ as the sources. After updating costs on all nodes in $\mathcal{G}$, the shortest path can be traced back from a node in $\mathcal{G}_M$ having the smallest cost. We start from the sampling rate with $n=4$. If no path can be found on $\mathcal{G}$, we double the value of $n$ to generate a graph with denser nodes and search the path again. 


A shortest path that minimizes the total cost of transition $\mathbb{J}_{trans}$ does not directly lead to an optimized path with minimum jerk. However, it provides a good initial path to be further optimized. 
This algorithm for obtaining an initial trajectory is resolution complete. It means that the algorithm in finite time either finds a solution if one exists, or correctly reports failure. The failure case happens when there exists no path satisfying the velocity limit $\mathbf{v}_{\max}$. 

\subsection{Improvement by Local Filtering}\label{subsecGreedyAlg}
With a given sequence of configurations $\mathcal{Q}$ in joint space according to the waypoints in $\mathcal{X}$, the jerks of trajectory at these discrete time points can be evaluated by the method of local approximation -- the formulation will be given below. In general, the trajectory determined by computing the shortest path on the graph $\mathcal{G}$ in the initialization step may also lead to jerky motion as the initial path does not directly minimize the total jerk (i.e., $\mathbb{J}$ in Eq.(\ref{eqGlobalOptm})). To reduce the total jerk on a trajectory with large number of waypoints, we develop an algorithm to improve the trajectory in an iterative \revision{way.}{routine after obtaining an initial trajectory by the graph-search method presented above.}

\begin{algorithm}[t!]
  \caption{Jerk Optimization by Local Filtering}\label{algLocalFiltering}

  \LinesNumbered
  \KwIn{An initial joint trajectory $\mathcal{P}$.}
  \KwOut{An optimized trajectory $\mathcal{P}$.}
  
  Set all points in $\mathcal{P}$ as \textit{free};
  
  Find a point $\mathbf{q}_c$ with maximal jerk on $\mathcal{P}$;
  
  \While{$|\dddot{\mathbf{q}}_c| > \mathbf{j}_{\max}$ \textbf{\textsc{AND}} not enough iterations}{
  
    Build a local path $\Tilde{\mathcal{P}}$ centered at $\mathbf{q}_c$ with margin $d$;
    
    $success$ = $false$;

    \While{$success \neq true$ \textbf{\textsc{AND}} $d \leq d_{\max}$}{
    
        $success$ = \textit{Minimize}~$(\mathbb{J}_{local}(\Tilde{\mathcal{P}}))$;
    
        \If{$success$}{
            Mark all points in $\Tilde{\mathcal{P}}$ as \textit{free};
        }
        \Else{
            Expand $\Tilde{\mathcal{P}}$ centered at $\mathbf{q}_c$ by $d=d+5$;    
        }
    }
    
    \If{$success \neq true$}{
         Mark all points in $\Tilde{\mathcal{P}}$ as \textit{locked};
    }
  
    Find $\mathbf{q}_c$ with maximal jerk only among the free points on $\mathcal{P}$;
  }
  
  \textbf{return} $\mathcal{P}$;

%
%
%
%
%
%
%
\end{algorithm}

Our algorithm is based on a greedy strategy. The pseudo-code of our algorithm is given in Algorithm \ref{algLocalFiltering}. At each iteration, we choose a trajectory point $\mathbf{q}_c$ with the maximum jerk among all points as $c=\arg \max_i \{ \|\dddot{\mathbf{q}_i}\|_W \}$. A local path $\Tilde{\mathcal{P}}$ is extracted with $\mathbf{q}_c$ as the center by using a margin $d$ (i.e., $\Tilde{\mathcal{P}}=\{\mathbf{q}_{a},\cdots,\mathbf{q}_{b}\}$ with $b,a=c \pm d$). We apply the local filtering to optimize the local path $\Tilde{\mathcal{P}}$ at the jerk level by solving the total-jerk problem defined on this local region (as $\mathbb{J}_{local}$ in Eq.(\ref{eqLocalOptm})). The following \revision{design of}{} algorithm is developed to further enhance the capability in jerk optimization. 
\begin{itemize}
\item \textit{Window-Size Adaptation}: When no feasible solution is found -- i.e., the value of $\mathbb{J}_{local}$ cannot be reduced, it means the computation of optimization is stuck at a local minimum. We then enlarge the window-size of filtering by including more trajectory points until it reaches a user-specified bound. 

\item \textit{Locking Mechanism}: A locking mechanism is developed in our algorithm to further enhance its capability to overcome the local optimum. Specifically, when the window-size of a local path has reached its maximally allowed bound, we will mark all the points in this path as \textit{locked}. The locked points will not be included in the further selection of points with maximal jerk (i.e., the center of local path extraction). A point will be \textit{unlocked} if it has been covered by some other local path, the locally defined total-jerk of which can be reduced. 
\end{itemize}
The adaptive local filtering with locking mechanism is repeatedly applied to regions with maximal jerk until the trajectory meets the required jerk-limit $\mathbf{j}_{\max}$ or the maximum number of iteration is reached (which however rarely occurs in our experiments). Note that, the requirement on maximal jerk is achieved by the algorithm instead of the numerical optimization conducted during the local filtering. 

Details for evaluating derivatives at waypoints and computing local-filter at the jerk level are presented as follows.

\subsubsection{Derivatives at Waypoints}\label{subsubsecDerivatives} 
To compute the derivatives of $\mathbf{q}$, we construct local curves interpolating the waypoints. Specifically, we have
\begin{equation}\label{eqLocalCurve}
\mathbf{q}_i(t) = \sum_{j=-2}^{2} B_j(t) \mathbf{q}_{i+j}
\end{equation}
with the basis functions $B_j(t)$ determined by imposing the interpolation constraints: $\mathbf{q}_i (t_{i+j})= \mathbf{q}_{i+j}$ ($\forall j=-2,\cdots,+2$). This results in
\begin{equation}\label{eqBasisFunc}
B_j(t)=\sum_{k=0}^4 b_{k,j+2} t^k,
\end{equation}
where $b_{k,j+2}=\beta_{k+1,j+3}$ with
\begin{equation}
 [\beta_{a, b}]_{5 \times 5} =
\begin{bmatrix}
    1  & t_{i-2} & t_{i-2}^2 & \dots & t_{i-2}^4 \\
    1  & t_{i-1} & t_{i-1}^2 & \dots & t_{i-1}^4 \\
    \vdots  & \vdots & \vdots & \ddots & \vdots \\
    1  & t_{i+2} & t_{i+2}^2 & \dots & t_{i+2}^4
\end{bmatrix}^{-1}.
\end{equation}
As a result, $\dddot{\mathbf{q}}(t_i)$ can be approximated as:
\begin{equation}\label{eqJerk}
\dddot{\mathbf{q}}(t_i) = \sum_{j=-2}^{2} \dddot{B_j}(t_i) \mathbf{q}_{i+j}.
\end{equation}
When having a constant time-interval $h = t_{i+1}-t_i$ between all waypoints and letting $t_i=0$, we will have a very compact formula in this special case:
\begin{equation}
\dddot{\mathbf{q}}(t_i) = \frac{\mathbf{q}_{i+2}-2\mathbf{q}_{i+1}+2\mathbf{q}_{i-1}-\mathbf{q}_{i-2}}{2 h^3},
\end{equation}
This is in fact the central finite difference formula for the third order numerical derivative obtained by applying the Taylor expansion. 

\subsubsection{Filter at the Jerk Level} 
Our filter is applied to a sequence of joint configurations as $\{\mathbf{q}_a, \mathbf{q}_{a+1}, \cdots, \mathbf{q}_b\}$ to minimize the jerk while still satisfying the discrete time constraints at $\{\mathbf{x}_a, \mathbf{x}_{a+1}, \cdots, \mathbf{x}_b\}$. We formulate the filter as a local optimization problem to minimize the sum of jerks at the waypoints.
\begin{equation}\label{eqLocalOptm}
\begin{aligned}
            & \min_{\{ \mathbf{q}_a, \cdots, \mathbf{q}_b\}} \mathbb{J}_{local}=\sum_{i=a}^b \| \dddot{\mathbf{q}}(t_i) \|_W^2 \\
s.t. \quad  & \mathbf{x}(t_i)=f(\mathbf{q}(t_i)) \quad (\forall i=a,\ldots,b), \\
            & \Gamma(\mathbf{q}(t_i))<0, \\
            & \mathbf{q}_{\min} \leq \mathbf{q}(t_i) \leq \mathbf{q}_{\max}, \\ 
            & |\dot{\mathbf{q}}(t_i)| \leq \mathbf{v}_{\max}, \, |\ddot{\mathbf{q}}(t_i)| \leq \mathbf{a}_{\max}. 
\end{aligned}
\end{equation}
The optimization problem defined in Eq.(\ref{eqLocalOptm}) is a non-convex problem with non-linear constraints. We therefore use \textit{sequential quadratic programming} (SQP) to solve it. When evaluating $\dddot{\mathbf{q}}_a$ and $\dddot{\mathbf{q}}_b$, those waypoints located at \revision{}{the} margin (i.e., $\dddot{\mathbf{q}}_{a-2}$, $\dddot{\mathbf{q}}_{a-1}$, $\dddot{\mathbf{q}}_{b+1}$ and $\dddot{\mathbf{q}}_{b+2}$) are involved as constants instead of variables. Therefore, when applying this local filter to different regions of the path $\mathcal{P}$, we should reserve a margin with at least \textit{four} points between the regions that will be locally updated. Note that we do not impose the requirement of maximally allowed jerk $\mathbf{j}_{\max}$ in this local filter as it is considered at the algorithm level.

\begin{figure}[t]
\centering
\includegraphics[width=\linewidth]{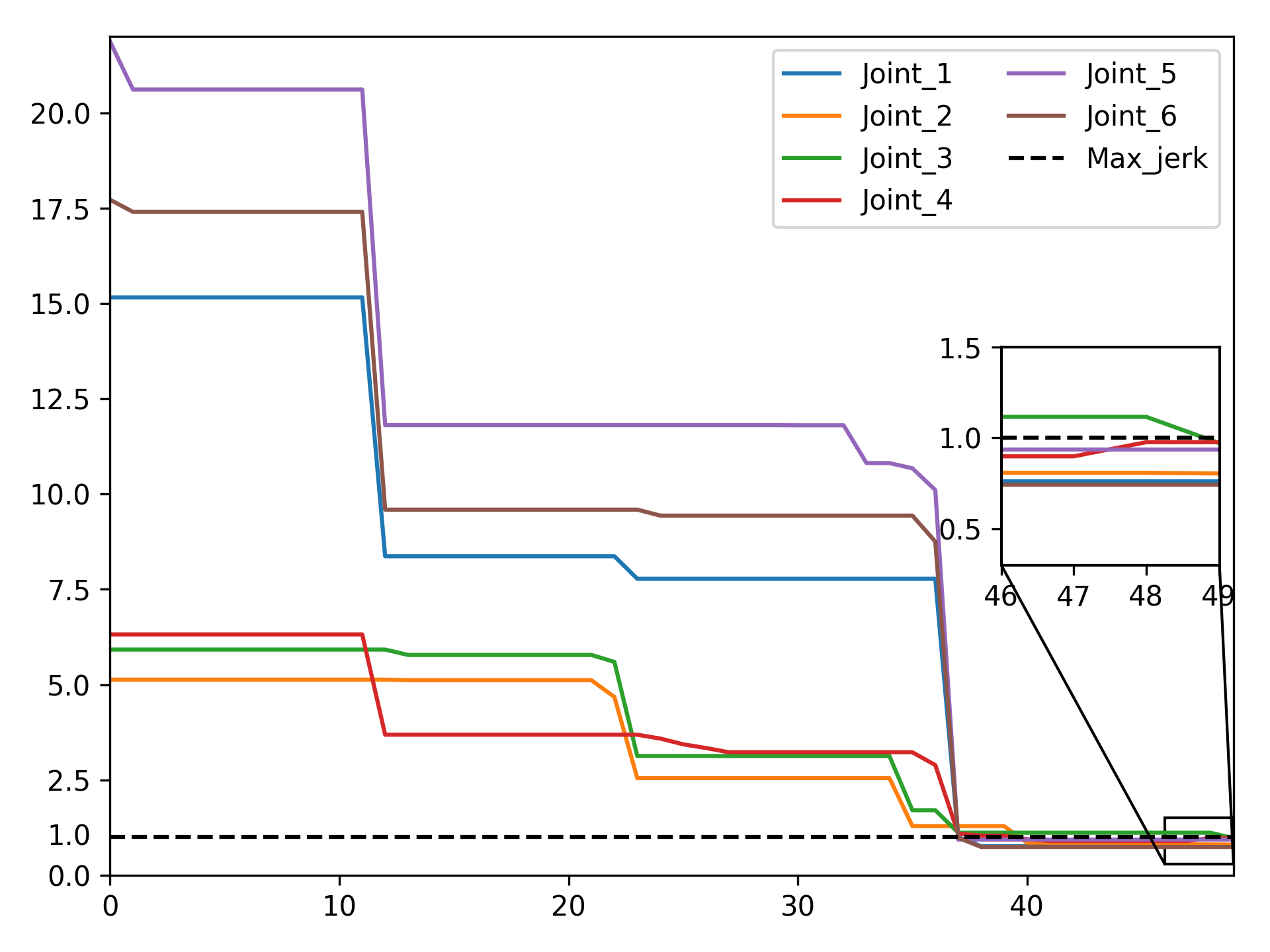}
\caption{The change of the maximum jerk at each joint during the iterations of our method. The maximal jerk has been reduced by $83.6\%-95.8\%$ on all the six joints. The dash line shows the allowed maximal jerk as $j_{\max}=1.0$ in this example.}
\label{figMaxJointJerk}
\end{figure}
\begin{figure}[t]
\centering
\includegraphics[width=\linewidth]{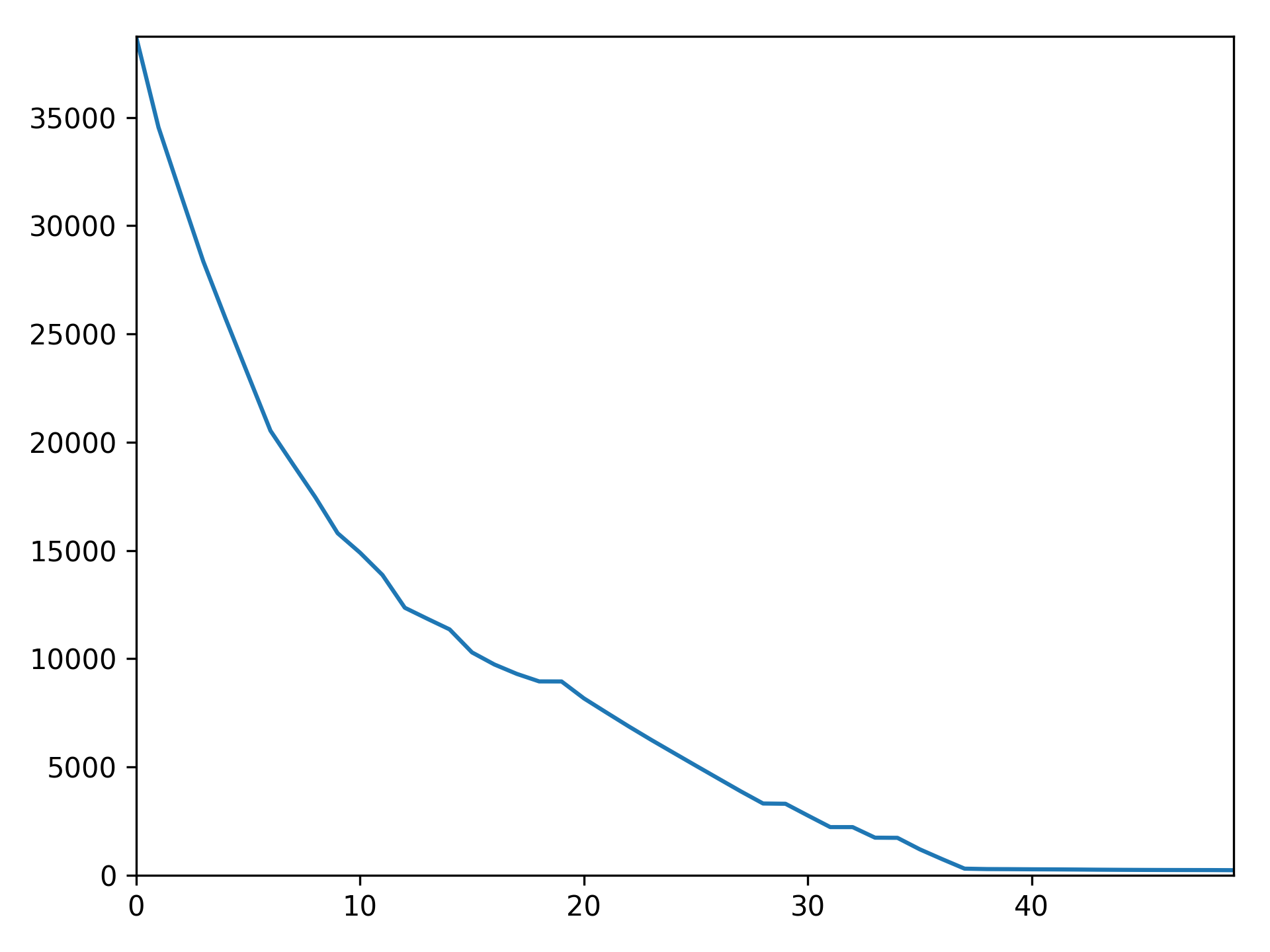}
\caption{The total sum of squared jerks, $\mathbb{J}$ in Eq.(\ref{eqGlobalOptm}), on the trajectory is effectively reduced during the iterations of our method. The value has been reduced by $99.4\%$ on the final result.
}
\label{figTotalJointJerk}
\end{figure}

\vspace{6pt}
The maximally allowed number of iterations in our greedy algorithm is set as $100$ in the implementation. Other parameters are set as $d=5$ and $d_{\max}=20$ according to our experiments. As shown in Figs.~\ref{figMaxJointJerk} and \ref{figTotalJointJerk}, our algorithm for trajectory optimization can effectively and efficiently reduce both the total jerk $\mathbb{J}$ and the maximal jerk on initial trajectories determined by the graph search. 

\subsection{Collision-free Verification}\label{subsecCollisionVerif}
%
After computing an optimized path $\mathcal{P}$, we need to verify the collision-free at every configurations on the path. This is implemented by applying an advanced collision-detection library (e.g., the FCL library \cite{Pan12}) at every point $\mathbf{q}_i \in \mathcal{P}$. 

If collision is found at $\mathbf{q}_d$, we adopt the following projection method to correct it:
\begin{itemize}
\item Re-sampling the rotational angle $\theta$ around the corresponding waypoint $\mathbf{x}_d$ in a very dense rate (i.e., with the step of $\pi/500$);

\item computing the collision-free IK solutions $\{ \hat{\mathbf{q}}_d^k \}$ for these dense samples;

\item selecting the closest one to serve as a projected solution as:
\begin{equation}
    \mathbf{q}_d = \arg \min_{\{ \hat{\mathbf{q}}_d^k \}} \| \mathbf{q}_d - \hat{\mathbf{q}}_d^k \|_{\infty}.
\end{equation}
\end{itemize}
The infinity norm is employed here to control the maximal variation for all joints. The result of projection may still violate the requirement of maximal jerk. Fortunately, our approximation of $\Gamma$ is accurate (see next Section) and we did not observe such a scenario in our experiments. In the worse case, we can still split one trajectory into two short trajectories to avoid large jerk in motion.

%% file: secCollisionClassifierNew.tex
\section{Learning-Based Collision Estimation}\label{secCollision}
An efficient method for estimating the collision-indication function $\Gamma(\cdot)$ is needed for solving the jerk-minimization problem by numerical optimization. Collision checking is a computational bottleneck for motion planning. For the general shape of obstacles, there is no trivial mapping from the working space to the configuration space. For the sake of efficiency, we develop a sampling-based approach to learn a function $\Tilde{\Gamma}(\cdot)$ to accurately approximate the collision-indication function $\Gamma(\cdot)$. Without loss of generality, collision detection can be formulated as a binary classification problem with $\Tilde{\Gamma}(\cdot)<0$ for collision-free and $\Tilde{\Gamma}(\cdot) \geq 0$ for collided configurations. This section first introduces our machine learning method for the representation of $\Tilde{\Gamma}(\cdot)$, and then introduces our \revision{manifold-centered}{contact centered} sampling strategy used to reduce the required number of training samples.

\subsection{Approximate Representation of Collision Function}\label{subsecCollisionRep}
Kernel-based function representations such as \textit{support vector machines} (SVM) and neural networks can be used to generate an algebraic function for $\Tilde{\Gamma}(\cdot)$. In our work, we choose SVM as: 1) learning of SVM is a convex optimization problem that can be efficiently solved, and 2) SVMs yields sparser models for high-dimensional non-linear classification problems -- i.e., with less number of kernel functions so that the value of $\Tilde{\Gamma}(\cdot)$ can be evaluated more efficiently.

Briefly, a SVM algorithm learns a hypothesis function which maps data from an input space to the feature space. Here, the input space is the configuration space $\{ \mathbf{q} \}$ and the feature space is the status of collision. Given $n$ sample points with labels obtained by the geometry-based collision detection library, we can learn a \textit{radial basis function} (RBF) based representation of $\Tilde{\Gamma}(\cdot)$ as
\begin{equation}\label{eqRadius}
\Tilde{\Gamma}(\mathbf{q})=\sum_{i=1}^{N} \alpha_i K(\mathbf{q}_i,\mathbf{q}) + b
\end{equation}
by using the Gaussian kernel function
\begin{center}
$K(\mathbf{q}_i,\mathbf{q}) = \exp{(-\gamma \| \mathbf{q} - \mathbf{q}_i \|^2)}$.
\end{center}
The learning result is the centers of kernels $\{ \mathbf{q}_i \}$ as the sub-set of training samples, the coefficients of RBFs $\{ \alpha_i \}$ and the value of a bias term $b$. As an algebraic function is provided here, we can evaluate the gradient of $\Tilde{\Gamma}(\mathbf{q}) \approx \Gamma(\mathbf{q})$ by the method presented in \cite{Seyed18} when solving the problem defined in Eq.(\ref{eqLocalOptm}). 

By the property of sparsity in SVM learning, the number of kernels $N$ could be much less than the number of training samples $n$. It is desirable to obtain fewer kernels so that $\Tilde{\Gamma}(\cdot)$ can then be evaluated more efficiently. $\gamma$ is a parameter specifying the narrowness of the Gaussian, and we use $\gamma=0.7$ in all our tests. Details of SVM learning and the method for tuning the parameter $\gamma$ can be found in \cite{Burges1998SVM,Scholkopf1997SVM}. 
Note that, to make the collision-indication provided by $\Tilde{\Gamma}(\cdot)$ \revision{be}{} more conservative, we update the value of the bias term $b$ \revision{as}{to} $b=b+\epsilon$ after obtaining the solution of SVM learning as $\Tilde{\Gamma}(\cdot)$. $\epsilon=1.0$ is employed in our implementation.

In the applications of robot-assisted manufacturing, the collision-indication function needs to be evaluated and also trained efficiently. The efficient evaluation is demanded as the function is used in the loop of the numerical optimization. The efficient training is also very important as the shape of a workpiece under fabrication (also considered as obstacles) is changed from time to time. By using the routine developed in \cite{dai18} for additive manufacturing, the collision-indication function needs to be updated when the fabrication process moves from one working surface layer to the next one. Similar concept of working surface layers can also be found in the subtractive manufacturing~\cite{Chen2018}. In general, we need to train a function $\Tilde{\Gamma}(\cdot)$ for each working surface when conducting the robot-assisted manufacturing. To reduce the training time, an effective way is to use less number of training samples. We develop a special sampling strategy for this purpose below. Our method can construct the training data-set, which is more capable to identify the boundary between collided and collision-free regions in the configuration space.

\subsection{Sampling Strategy for Training}\label{subsecSampling}
Inspired by the active learning method with a coarse-to-fine iterative sampling refinement strategy presented in \cite{Pan13}, we first generate sparse samples in the configuration space to capture the large scale topology of the indication function. This function is later refined by adding more selected samples near the decision boundary. 

\subsubsection{\revision{Contact-Manifold}{C-space of contact}}
To generate more effective training samples, we introduce a concept of \revision{contact-manifold}{contact \textit{configuration space} (C-space)} as a set of all configurations where the robotic system's tool touches a working surface $\mathcal{S}$. Given a forward kinematic function $\mathbf{f}(\cdot)$ of the robotic system, the \revision{contact-manifold}{contact C-space} of the working surface $\mathcal{S}$ is defined as:
\begin{center} 
$\mathcal{Q}_{cont}=\{ \mathbf{q} \; | \; dist(\mathbf{f}(\mathbf{q}),\mathcal{S})=0, \forall \mathbf{q} \in \mathbb{R}^L\ \}$ 
\end{center}
with $dist(\cdots)$ being the distance function. 

Samples are generated around $\mathcal{Q}_{cont}$ by the steps of initialization and the \revision{manifold-centered}{contact centered} refinement as presented below. A projection operator $\Upsilon(\cdot)$ is developed to project a general configuration $\mathbf{q}$ onto the \revision{contact-manifold}{C-space of contact} by solving the following minimization problem as 
\begin{equation}\label{eqProjection}
\begin{aligned}
    & \Upsilon(\mathbf{q}) = \arg \min_{ \mathbf{q}^* } \| \mathbf{q}^* - \mathbf{q} \|_2^2 \\
s.t. \quad  & dist(\mathbf{f}(\mathbf{q}^*),\mathcal{S})=0. 
\end{aligned}
\end{equation}
In our implementation, the solution of $\mathbf{q}^*$ is computed by the sequential quadratic programming and the distance function is efficiently evaluated by the \textit{Proximity Query Package} (PQP) library~\cite{Gottschalk96}.

\subsubsection{Initial sampling}\label{subsubInitSampleGen}
Sparse samples in $\mathcal{Q}_{cont}$ are initially generated by sampling the working surface $\mathcal{S}$. Specifically, we randomly sample $m$ points on $\mathcal{S}$. At each sample point, we can generate a quaternion by using the surface normal and a random angle $\theta$ as mentioned in Eq.(\ref{eqQuaternion}). The corresponding configuration of each quaternion can be obtained by the IK calculation, and the collision status is obtained by the geometry-based collision detection (e.g., the \textit{Flexiable Collision Library} (FCL) presented in \cite{Pan12}). 
This gives the initial set of training samples -- all \revision{on contact-manifold}{from the C-space of contact}.

\subsubsection{Up-scaling and Refinement}
Two steps are employed to generate more samples in the nearby region around the \revision{contact-manifold}{the contact C-space} $\mathcal{Q}_{cont}$. 
\begin{itemize}
\item 
\textbf{Up-scaling:} The purpose of this step is to generate nearly uniform samples in the nearby region of $\mathcal{Q}_{cont}$ for capturing the topological structure of the collision-indication function $\Gamma(\cdot)$.
Randomly sampling the working surface will not enable this uniformity in the configuration space as the mapping of IK can be very complicated. Directly generating random samples in the configuration space would however require much more samples to capture the structure: most of samples generated in this way would be far away from the \revision{contact-manifold}{C-space of contact,} $\mathcal{Q}_{cont}$. 
%
%
Differently, we generate more samples near the \revision{contact-manifold}{C-space of contact} by up-sampling the initial set of samples. 
When the distance between an existing sample and any of its $k$-nearest neighbors is larger than a threshold $\tau_{\alpha}$, a new sample is generated in the middle. {In effect,} $50\%$ of the newly generated samples will be projected onto the \revision{contact-manifold}{C-space of contact,} $\mathcal{Q}_{cont}$. The up-scaling is repeated until no new sample can be generated under the density control of $\tau_{\alpha}$ (e.g., $\tau_{\alpha}=0.8$ is chosen in our implementation by empirical tests).

\item \textbf{Refinement:} After up-scaling, a step of boundary-aware refinement is applied to generate samples for learning a more precise decision boundary when approximating the collision-indication function $\Gamma(\cdot)$. Similarly, this is based on searching the $k$-nearest neighbors of existing samples. When a sample and its neighbor have different collision labels, we generate a new sample if their distance is larger than a threshold $\tau_{\beta}$. A denser sampling is desired along the boundary; therefore, $\tau_{\beta}<\tau_{\alpha}$ is used (e.g., $\tau_{\beta}=0.05$ in our implementation). Again, the $50\%$ of the newly generated samples will be projected onto $\mathcal{Q}_{cont}$, and the refinement is repeatedly applied until no new sample can be generated under the density control of $\tau_{\beta}$.
\end{itemize}
These two steps are repeatedly applied until the specified total number of samples has been generated. For the nearest neighbors search, $k=20$ is used in all our examples. The pseudo-code of these two steps can be found in Algorithm \ref{algSampling}. As can be found in the following sub-section of analysis, the approximation function generated by SVM can better capture the indication function with the help of much smaller number of training samples. 

\begin{figure}
\centering
\includegraphics[width=\linewidth]{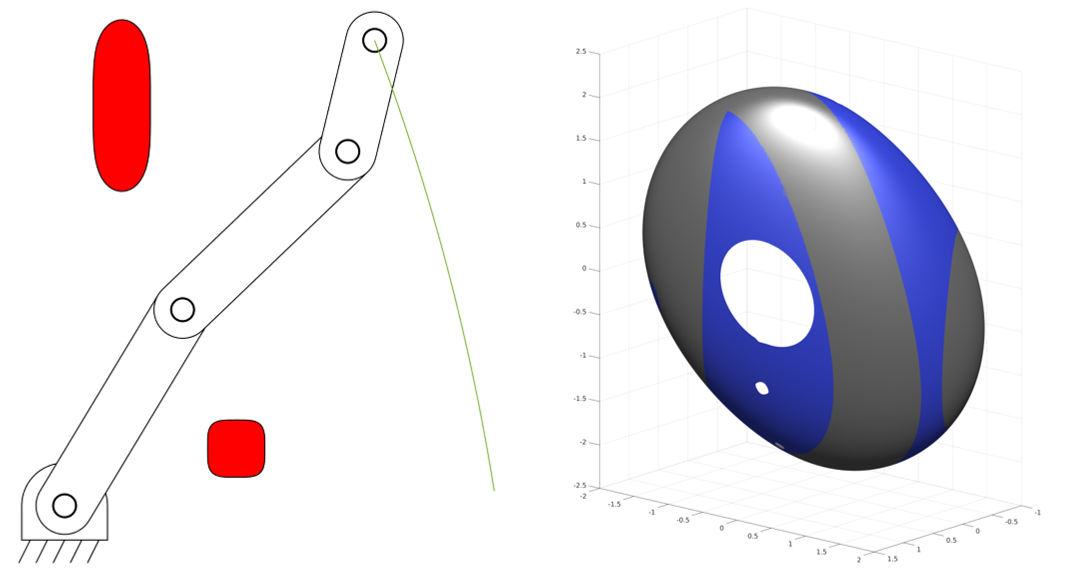}
\caption{A 3-DOF planar robotic arm for tracing a 2D path (green) with obstacles (red). The example is used to study the effectiveness of our sampling strategy for learning the collision-indication function. As shown in the right, the \revision{contact-manifold}{C-space of contact} $\mathcal{Q}_{cont}$ is displayed by blue color for the collision-free region (i.e., $\mathcal{Q}_{cont} \cap \mathcal{Q}_{free}$) and gray color for the collided region (i.e., $\mathcal{Q}_{cont} \cap \overline{\mathcal{Q}_{free}}$). Note that, the white regions in the configuration space are not reachable by the robotic arm.}
\label{figPlanarExample}
\end{figure}
\begin{figure*}[t]
\centering
\includegraphics[width=\linewidth]{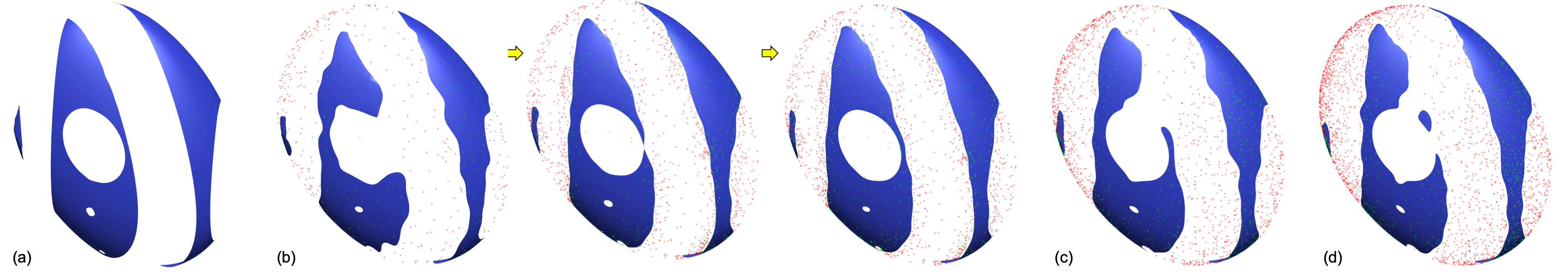}
\caption{Comparisons to demonstrate the effectiveness by using the samples generated by our method to learn a function $\Tilde{\Gamma}(\cdot)$ to approximate the collision-indication function $\Gamma(\cdot)$ (a). From the left to right in (b), the progressive sampling results and their corresponding $\Tilde{\Gamma}(\cdot)$ are obtained by SVM are shown as: i) after the first round of up-scaling (with $n=474$ and $N=348$), ii) after the first round of refinement (with $n=1527$ and $N=714$), and iii) the final result of sample generation (with $n=1698$ and $N=779$) after a few iterations. The learning results are worse than ours if random sampling is employed to generate (c) the same number of samples as ours and (d) the same number of kernels as ours, where (c) $1698$ samples only can result in $517$ effective kernels and (d) needs $2553$ samples to obtain $779$ effective kernels. 
}\label{figSamplingComparison}
\end{figure*}

\begin{algorithm}[h!]
\caption{Generate Samples for Training}\label{algSampling}

\LinesNumbered
\KwIn{A working surface $S$.}
\KwOut{A resultant set $\mathcal{Q}$ of samples.}

\tcc{The step of \textit{Initialization}}
  
Randomly generate $m$ samples for $\mathcal{Q}$ with all on $\mathcal{Q}_{cont}$;
  
  \Repeat{no new point can be added into $\mathcal{Q}$}{
      \tcc{The step of \textit{Up-scaling}}    
      \ForEach{$\mathbf{q}\in \mathcal{Q}$}{
        \If{$\mathbf{q}\in \mathcal{Q}_{cont}$}{
            Search the $k$-NN of $\mathbf{q}$ as a set $\mathcal{Q}_{NN}$;
            
            \ForEach{$\mathbf{q}^*\in \mathcal{Q}_{NN}$}{
                \If{$\| \mathbf{q}^* - \mathbf{q}\|>\tau_{\alpha}$}{
                    \textit{GenerateNewSample}($\mathbf{q}$, $\mathbf{q}^*$, $\mathcal{Q}$);
                }
            }
        }
      }

  \tcc{The step of \textit{Refinement}}    
      \ForEach{$\mathbf{q}\in \mathcal{Q}$}{
        \If{$\mathbf{q}\in \mathcal{Q}_{cont}$}{
            Search the $k$-NN of $\mathbf{q}$ as a set $\mathcal{Q}_{NN}$;
            
            \For{every $\mathbf{q}^*\in \mathcal{Q}_{NN}$}{
                \If{$\| \mathbf{q}^* - \mathbf{q}\|>\tau_{\beta}$ \textbf{\textsc{AND}} $L(\mathbf{q}^*) \neq L(\mathbf{q})$}{
                    \tcc{$L(\cdot)$ returns the collision status}
                
                    \textit{GenerateNewSample}($\mathbf{q}$, $\mathbf{q}^*$, $\mathcal{Q}$);
                }
            }
        }
      }
  }

\textbf{return} $\mathcal{Q}$;

\BlankLine

\textbf{Procedure}~\textit{GenerateNewSample}($\mathbf{q}$, $\mathbf{q}^*$, $\mathcal{Q}$) 

\Begin{
$\mathbf{q}_{new}= \frac{1}{2}(\mathbf{q} + \mathbf{q}^*)$;

\tcc{50\% new samples projected}

Generate a random $\rho \in [0,1)$;
                    
\lIf{$\rho \geq 0.5$}{$\mathbf{q}_{new}=\Upsilon(\mathbf{q}_{new})$}                     
Add $\mathbf{q}_{new}$ into $\mathcal{Q}$;
}

\textbf{End Procedure}

\end{algorithm}

\subsection{Analysis}\label{subsecSamplingAnalysis}
We employ a 3-DOF planar redundant robotic arm as an example to analyze the effectiveness of our sampling strategy in a 2D path tracing setup (see the left of Fig.~\ref{figPlanarExample}). 
To follow the 2D path displayed in green color, the three joints of this robot form a \revision{contact-manifold}{C-space of contact} in the configuration space (see the right of Fig.~\ref{figPlanarExample}). When presenting obstacles as the red objects, the blue regions denote the collision-free configurations on the \revision{contact-manifold}{C-space of contact}. In this analysis, we study how our sampling-and-learning method can effectively capture the boundary between collision-free and collided regions. 

Figure \ref{figSamplingComparison} shows our results comparing to those of SVM-learning by random samples. The ground-truth collision-indication function on the \revision{contact-manifold}{C-space of contact} is shown in Fig.~\ref{figSamplingComparison}(a). The progressive results of our sample generation algorithm are given in Fig.~\ref{figSamplingComparison}(b), from which it is easy to find that samples generated by our method properly capture the boundary of the indication function after the steps of up-scaling and refinement. As can be seen in Fig.~\ref{figSamplingComparison}(c), the structure and the boundary of collision-free regions cannot be captured when the same number of samples are generated randomly. In this case, less effective kernels are obtained by SVM-learning. The region of collision-free configurations can be better captured when using more random samples -- see the result shown in Fig.~\ref{figSamplingComparison}(d), where the same number of effective kernels are obtained by SVM-learning. However, the function learned in this case is still less accurate than ours. In summary, the sampling strategy developed in our algorithm can better capture the boundary of an collision-indication function when SVM-learning is adopted. More experiments about the prediction rate and the checking time will be shown in the following section. 

%% file: secExperimentResult.tex
\begin{figure}[t]
\centering
\includegraphics[width=\linewidth]{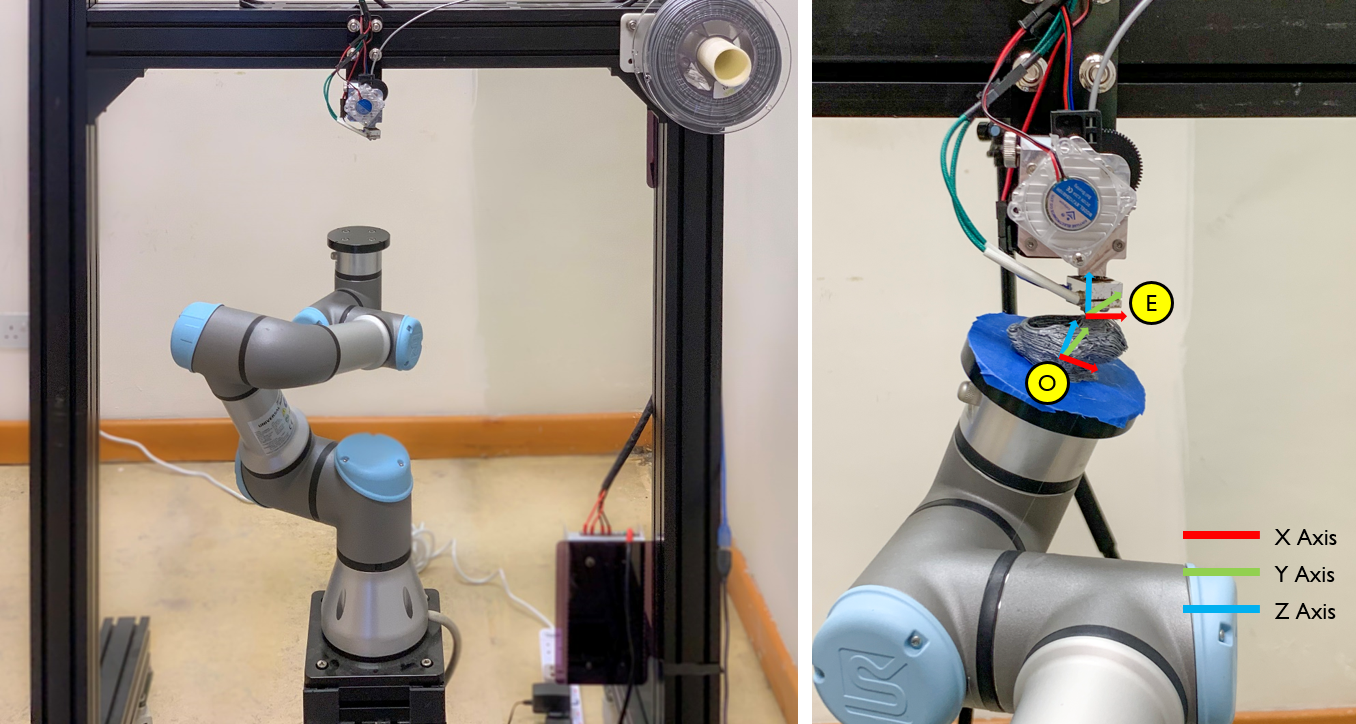}
\caption{A hardware setup of our robot-assisted 3D printing system with 6-DOF. (Left) The UR3 robot-arm based setup with a fixed material-extruder for better material adhesion. (Right) With the help of the relationship between the extruder frame $E$ and the frame of workpiece $O$, an analytic inverse kinematic solver can be employed to obtain configurations in the joint space from a quaternion determined by rotating the frame $E$ around the orientation given in a waypoint.
}\label{figHardware}
\end{figure}

\section{Experimental Results}\label{secResult}
\begin{figure}
\includegraphics[width=\linewidth]{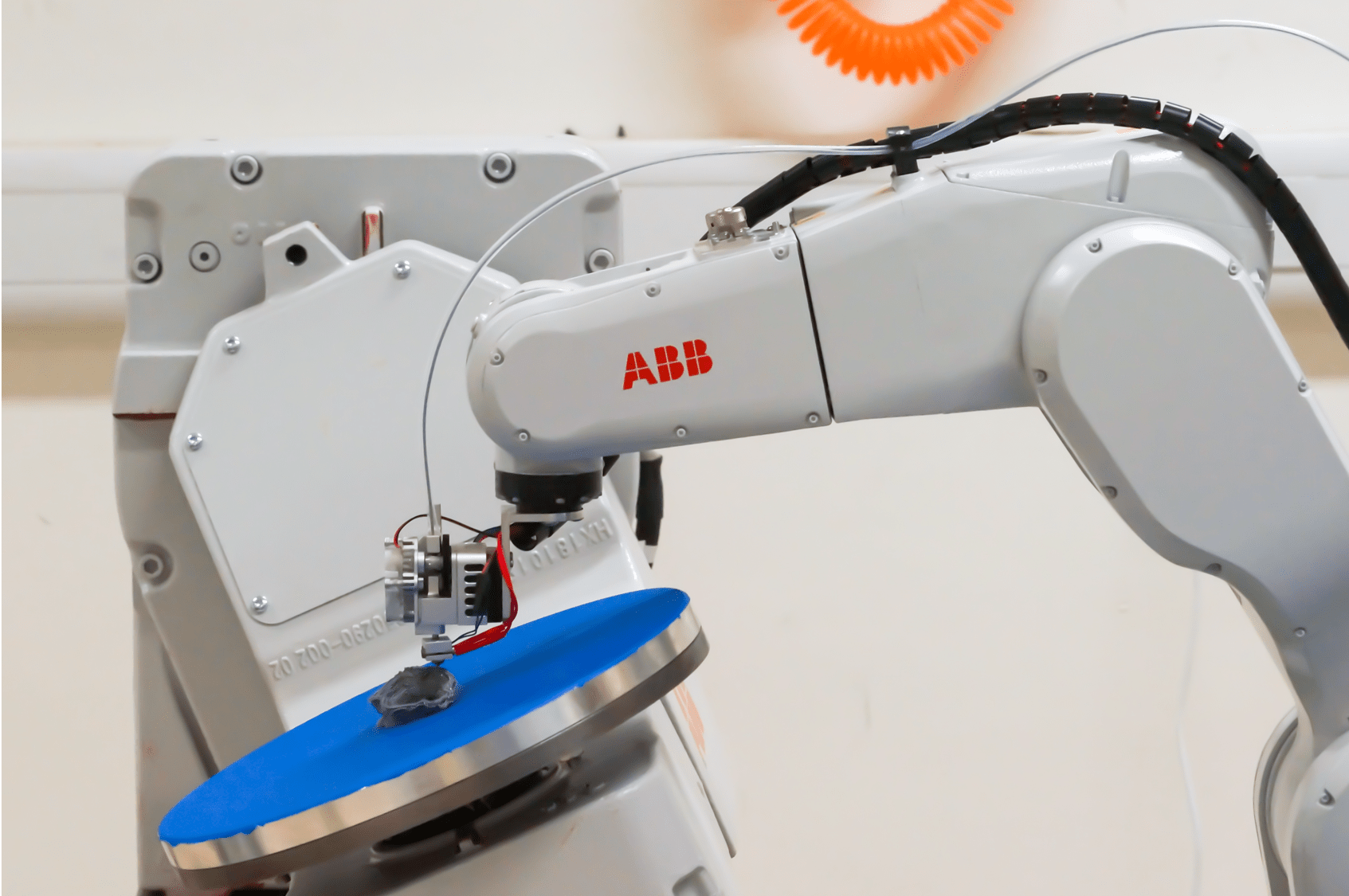}
\caption{A hardware setup of robot-assisted 3D printing system with 8-DOFs consisting of a 6-DOF ABB IRB1200-7/0.7 robotic arm and a 2-DOF IRBP A-250 tilting table.}\label{figABBHardware}
\end{figure}

We implement our algorithm on \textit{Robot Operating System} (ROS) framework with our UR3 based robotic fabrication setup by using C++. All evaluations are executed on a PC with Intel\textsuperscript \textregistered Core\texttrademark i7 processor, 32 GB RAM and GeForce GTX 2070 video card, running Ubuntu 16.04 (Xenial, 64-bits OS). Besides of computational experiments, the performance of our approach has been demonstrated on two different hardware systems for robot-assisted 3D printing (i.e., Fig.~\ref{figHardware} for a \revision{6-DOFs}{6-DOF} system and Fig.~\ref{figABBHardware} for a \revision{8-DOFs}{8-DOF} system). 

Our planning algorithm can effectively and efficiently compute a smooth and collision-free trajectory of redundant robot. The quality of fabrication can be significantly improved as the jerk has been optimized on the motion trajectories. More details can be found in the reported experimental tests below.

\subsection{Learning Results of Collision-Indication Functions}
In our implementation, the libSVM library \cite{CC01a} was used for SVM-learning. The effectiveness of our sample generation method for SVM-learning based estimation of collision-indication function has been demonstrated by a planar redundant robot in Section \ref{subsecSamplingAnalysis} above. Here we further study its performance in robot-assisted fabrication by using 3D tool-paths. To quantitatively measure the accuracy of $\Tilde{\Gamma}(\cdot)$ for approximating $\Gamma(\cdot)$, we evaluate the following \textit{true-negative-ratio} (TNR) metric based on samples of verification.
\begin{equation}\label{eqTNR}
\textsc{TNR}=\frac{\textsc{TN}}{\textsc{TN}+\textsc{FP}},
\end{equation}
where $\textsc{TN}$ (\textit{True Negative}) denotes the number of samples with $\Tilde{\Gamma}(\cdot)<0$ that are really collision-free while $\textsc{FP}$ (\textit{False Positive}) gives the number of samples with $\Tilde{\Gamma}(\cdot) \geq 0$ that however will \textit{not} lead to collision or contact. $\textsc{TNR}$ provides the percentage of samples that are truly collision-free among all samples with negative value returned by the indication function $\Tilde{\Gamma}(\cdot)$, the larger the better.
%
%

\begin{table}[t]
\centering
\caption{Statistics for the Accuracy of Collision-Indication Functions Generated by SVM-Learning}
\begin{tabular}{|c|c|c||c|c|}
\hline
\textbf{Verification Type} & \multicolumn{2}{c||}{Nearby Region} & \multicolumn{2}{c|}{C-space of Contact} \\
\hline
     & WS1 & WS2 &  WS1 & WS2 \\
TNR  & Fig.\ref{figWorkingSurfaces}(a) & Fig.\ref{figWorkingSurfaces}(b) &  Fig.\ref{figWorkingSurfaces}(a) & Fig.\ref{figWorkingSurfaces}(b) \\
\hline
\hline
Our Method &  &    &   &  \\
($n=3020$ \& $N=887$) & $0.963$ & $0.944$  & $0.969$ & $0.956$  \\
\hline
\hline
By Random Samples &  &    &   &  \\
($n=3020$ \& $N=479$) & $0.894$ & $0.876$  & $0.942$ & $0.923$  \\
\hline
By Random Samples &  &    &   &  \\
($n=5842$ \& $N=887$) & $0.889$ & $0.883$  & $0.951$ & $0.943$  \\
\hline
\end{tabular}
\label{tableClassifierVerification}
\end{table}

\begin{figure}[t]
\centering
\includegraphics[width=\linewidth]{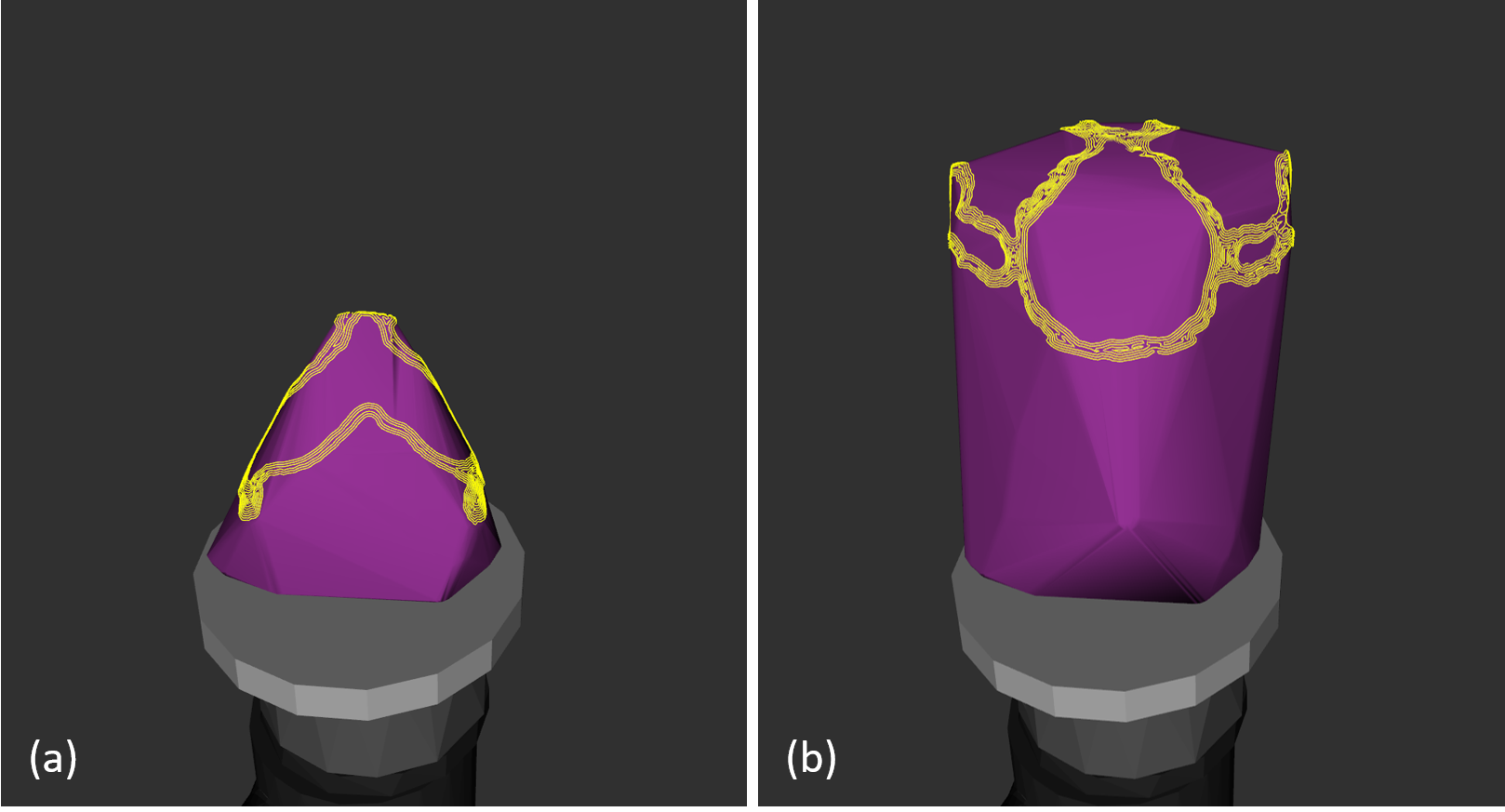}
\caption{Two working surfaces with tool-paths used in our experimental tests: (a) one working surface layer of the armadillo model with $2545$ waypoints -- its resultant trajectory is shown in Fig.~\ref{figResPathA} and (b) working surface layer of the armadillo model with $4,681$ waypoints having the optimized trajectory given in Fig.~\ref{figResPathB}.
}\label{figWorkingSurfaces}
\end{figure}

We generate two different types of samples to verify the accuracy of a classifier on different \textit{working surfaces} (WS) as shown in Fig.~\ref{figWorkingSurfaces}. In the first type, verification samples are generated at the nearby regions of the contact-manifold to simulate the situations while computing the numerical optimization. In the second type, verification samples are generated on the contact-manifold by the method of initial samples presented in Section \ref{subsecSampling}. In both types of tests, $100,000$ verification samples are employed for all examples. The true status of the samples are generated by the geometry-based collision-detection library. The resultant statistics can be found in Table \ref{tableClassifierVerification}. It is easy to find that the accuracy of our collision-indication function is much higher than the classifier generated by SVM-learning from random samples. Note that, for conducting a fair comparison we also provide the results of a random-sampling based classifier with the same number of kernels (i.e., the same value of $N$), which needs much more samples. The nearby sets of verification samples are more similar to the situation happens during numerical optimization. Comparing to the geometry computation based collision-detection technique (e.g., FCL library~\cite{Pan12}), the evaluation of $\Gamma(\cdot)$ with $N=887$ kernels is $5 \times$ faster if only \revision{collsion}{collision}-check is needed. However, the gradient of the collision-indication  function needs to be evaluated in the numerical optimization (i.e., Eq.(\ref{eqLocalOptm})). To provide a similar function by the geometry-based collision detection, we need to evaluate the distance to obstacles. In this case, our method is around $220\times$ faster.

\begin{figure*}[t]
\centering\includegraphics[width=\linewidth]{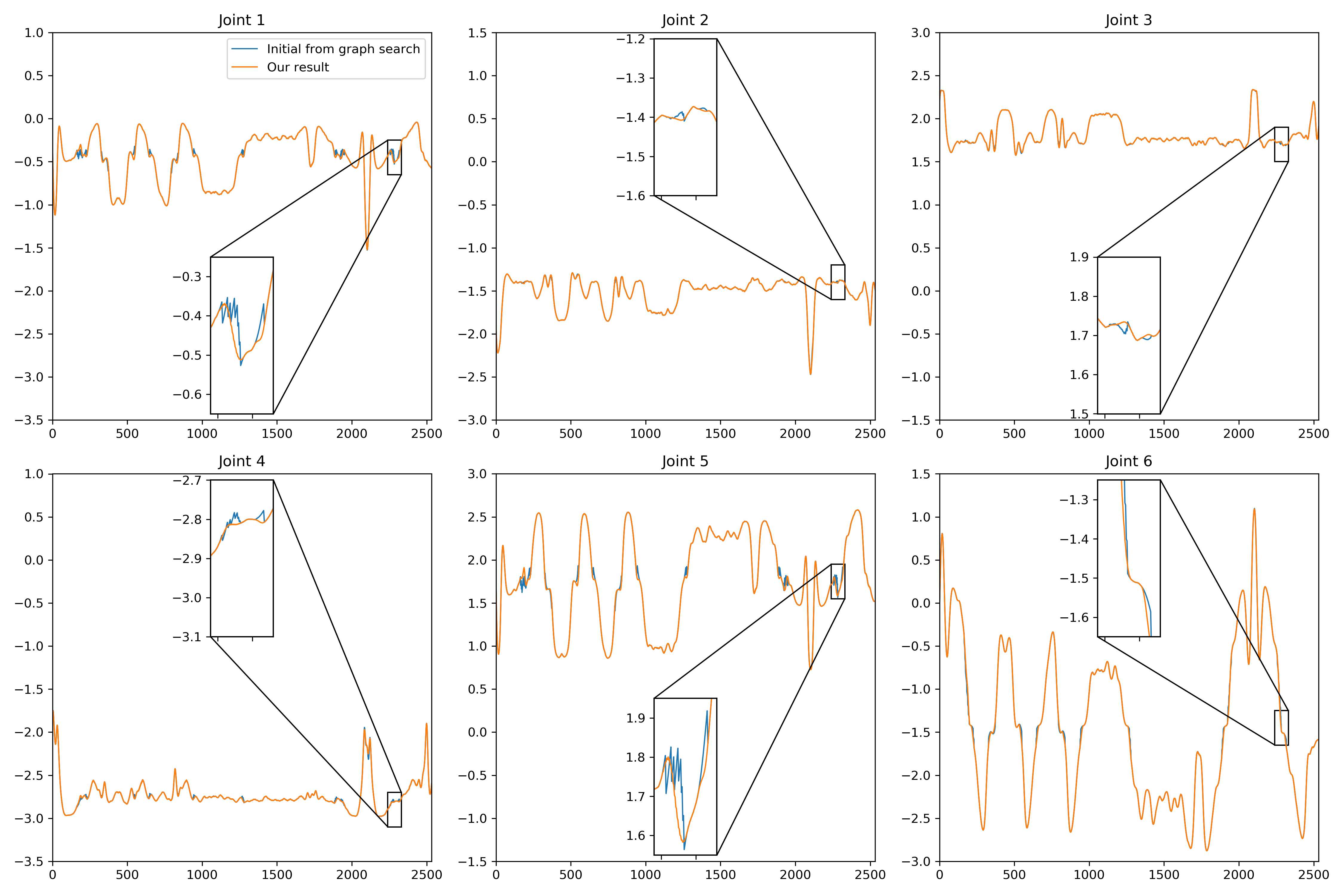}
\centering\caption{The comparison between the initial trajectory from graph-based search and after applying local refinement for each joint on the 6-DOF system. The zoom views shows clearly the smoothness improvement after the local refinement. The tool-path is as shown in Fig.~\ref{figWorkingSurfaces}(a).}\label{figResPathA}
\end{figure*}

\begin{figure}[t]
\centering\includegraphics[width=\linewidth]{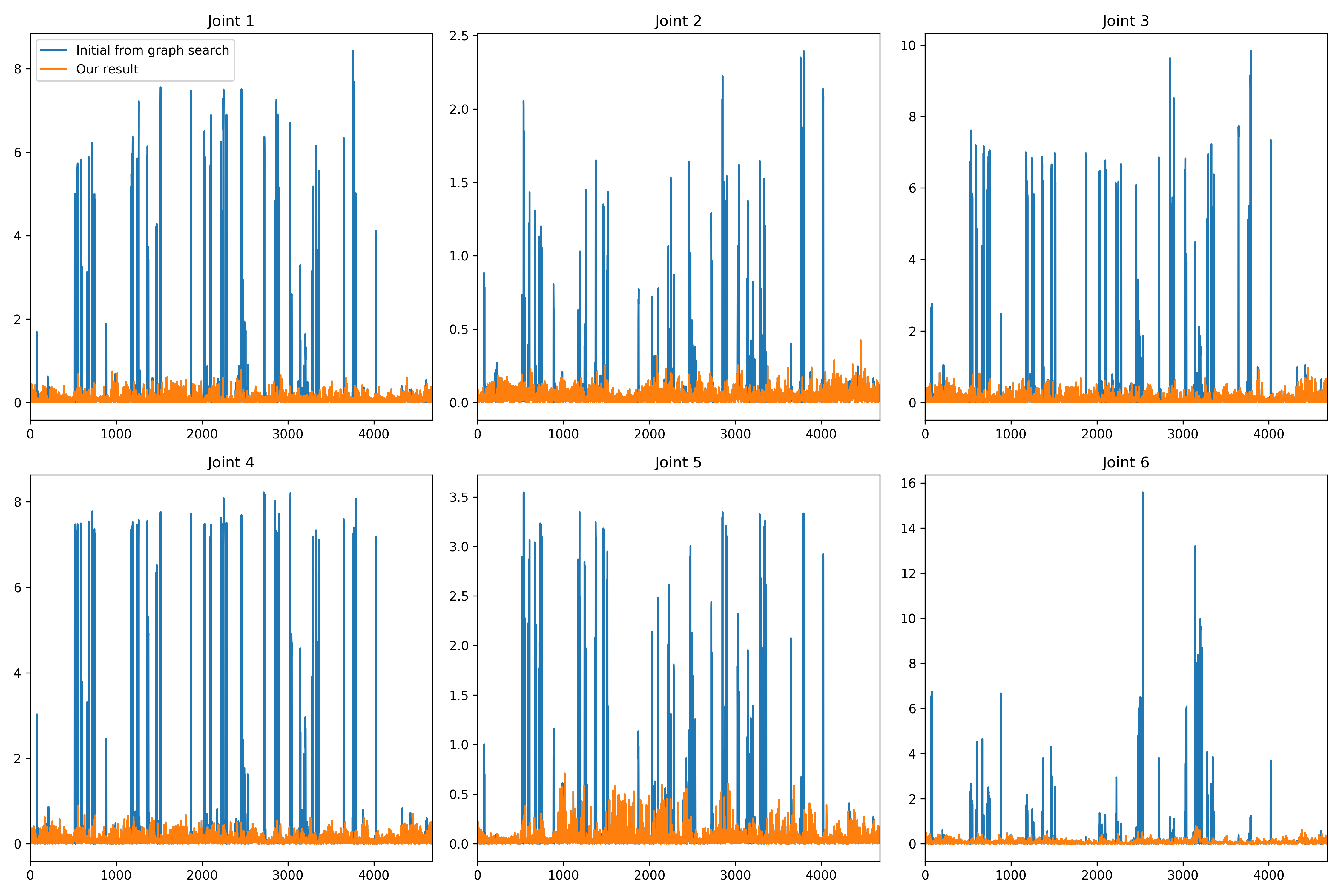}
\centering\caption{The comparison of jerk (the absolute value) between the initial trajectory from graph-based search (displayed in blue color) and after applying local refinement for each joint (displayed in orange color), where significant improvements can be easily observed. Again, this is implemented on the 6-DOF system, and the tool-path is as shown in Fig.~\ref{figWorkingSurfaces}(b).}\label{figResPathB}
\end{figure}

\begin{table}
\centering
\caption{Statistics of Computation}
\begin{tabular}{|c|c||c|c|c|c||c|c|}
\hline
\multicolumn{2}{|c||}{\textbf{Example}} &    \multicolumn{4}{c||}{\textbf{Our Method}}  &   \multicolumn{2}{c|}{\textbf{Dense Graph}} \\
\hline
\hline
    &  Wpt. &     \multicolumn{3}{c|}{Computing Time (sec.)}    &   $\mathbb{J}^*$ & Time &  $\mathbb{J}^*$  \\
\cline{3-5}
Path  &  Num.    & Init. &  Optm.  & Total & Eq.(\ref{eqGlobalOptm}) & (sec.) & Eq.(\ref{eqGlobalOptm}) \\
\hline
\hline
WS1  &  $2,545$  &  $86$   &   $30$ & $116$   & $0.232$  & $1,837$ & $1.34$ \\
WS2  &  $4,681$  &  $265$   &   $58$ & $322$   & $0.697$ & $3,283$ & $3.82$ \\
\hline
\hline
&    &  \multicolumn{4}{c||}{Maximal Jerk on Path}  &   \multicolumn{2}{c|}{Resultant}   \\
\cline{3-6}
 &  Joint  &  \multicolumn{2}{c|}{Before Optm.}   & \multicolumn{2}{c||}{After Optm.}  &   \multicolumn{2}{c|}{Maximal Jerk}   \\
\hline
WS1  &  1  & \multicolumn{2}{c|}{$15.16$} & \multicolumn{2}{c||}{$0.76$} & \multicolumn{2}{c|}{$1.85$} \\
&  2  & \multicolumn{2}{c|}{$5.13$} & \multicolumn{2}{c||}{$0.80$} & \multicolumn{2}{c|}{$1.09$}\\ 
&  3  & \multicolumn{2}{c|}{$5.92$} & \multicolumn{2}{c||}{$0.97$} & \multicolumn{2}{c|}{$1.75$}\\ 
&  4  & \multicolumn{2}{c|}{$6.32$} & \multicolumn{2}{c||}{$0.97$} & \multicolumn{2}{c|}{$1.35$}\\ 
&  5  & \multicolumn{2}{c|}{$21.87$} & \multicolumn{2}{c||}{$0.94$} & \multicolumn{2}{c|}{$2.24$}\\ 
&  6  & \multicolumn{2}{c|}{$17.72$} & \multicolumn{2}{c||}{$0.74$} & \multicolumn{2}{c|}{$5.26$}\\ 
\hline
WS2  &  1  & \multicolumn{2}{c|}{$8.43$} & \multicolumn{2}{c||}{$0.77$} & \multicolumn{2}{c|}{$2.23$} \\
&  2  & \multicolumn{2}{c|}{$2.39$} & \multicolumn{2}{c||}{$0.42$} & \multicolumn{2}{c|}{$2.55$}\\ 
&  3  & \multicolumn{2}{c|}{$9.83$} & \multicolumn{2}{c||}{$0.97$} & \multicolumn{2}{c|}{$4.18$}\\ 
&  4  & \multicolumn{2}{c|}{$8.22$} & \multicolumn{2}{c||}{$0.89$} & \multicolumn{2}{c|}{$1.58$}\\ 
&  5  & \multicolumn{2}{c|}{$3.55$} & \multicolumn{2}{c||}{$0.71$} & \multicolumn{2}{c|}{$1.98$}\\ 
&  6  & \multicolumn{2}{c|}{$15.59$} & \multicolumn{2}{c||}{$0.78$} & \multicolumn{2}{c|}{$3.25$}\\ 
\hline
\end{tabular}\label{tableCompStatistics}
\begin{flushleft}
$^*$The value of $\mathbb{J}$ is reported at the unit of $\times 10^3$.
\end{flushleft}
\end{table}

\subsection{Results of Jerk-optimized Trajectories}
In this sub-section, we show the resultant motion trajectories generated by our jerk-optimized planning method. The first example is a tool-path as shown in Fig.~\ref{figWorkingSurfaces}(a) for the 6-DOF robotic system. The progressive results for optimizing the trajectory have been shown in Figs.~\ref{figMaxJointJerk} and \ref{figTotalJointJerk}. It can be observed that our optimization approach can reduce both the maximal jerk and the total sum of squared jerk by $83.6\%-95.8\%$ and $99.4\%$ respectively. Fig.~\ref{figResPathA} shows the path's angular values on all six joints before and after the optimization, where the zoom views clearly show the improvement on smoothness.  
In the second example, the tool-path as shown in Fig.~\ref{figWorkingSurfaces}(b) is the target to be realized on the UR3-robot based hardware platform for 3D printing. The trajectories in joint space before and after optimization are given in Fig.~\ref{figResPathB}, where the maximal jerk on all joints have been reduced by 
up to $95\%$. 

We also compare the results generated by our method to the graph-search based method with a denser sampling -- i.e., the nodes in a ladder are generated by every degree for the value of $\theta$ in Eq.(\ref{eqQuaternion}). This is actually the method used in \cite{dai18}. Detail computational statistics of our trajectory planning algorithm in the robot-assisted 3D printing application can be found in Table \ref{tableCompStatistics}. It can be observed that our method generates a trajectory with much lower jerks (both the maximal jerk and the total jerks) while being $10\times$ faster. This demonstrates both the effectiveness and the efficiency of our approach. 

\subsection{Robot-Assisted 3D Printing}
\begin{figure}[t]
\includegraphics[width=\linewidth]{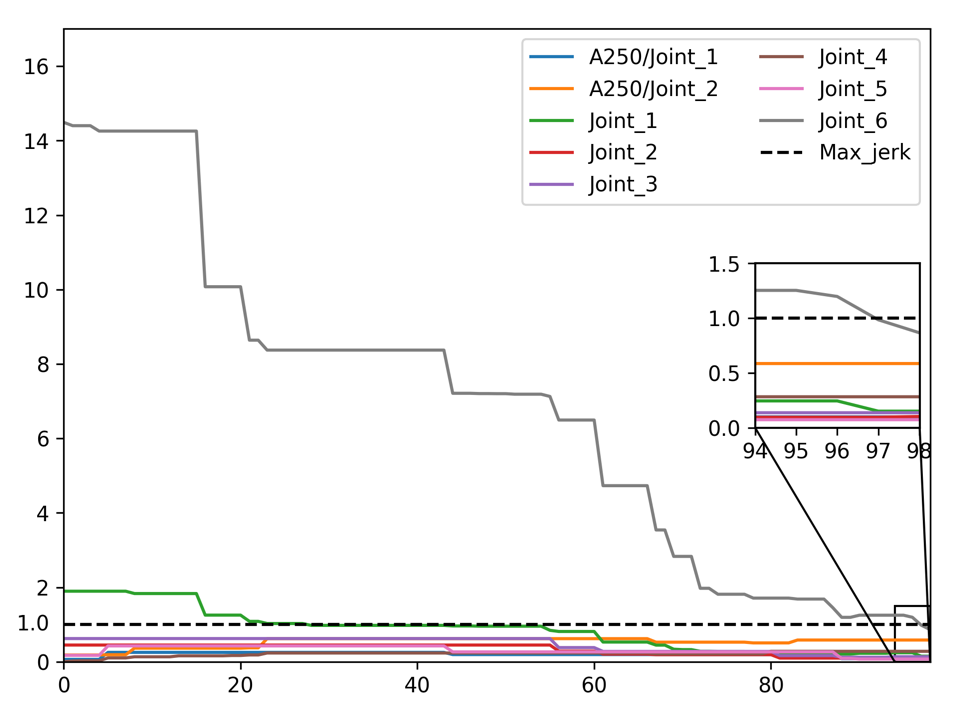}
\includegraphics[width=\linewidth]{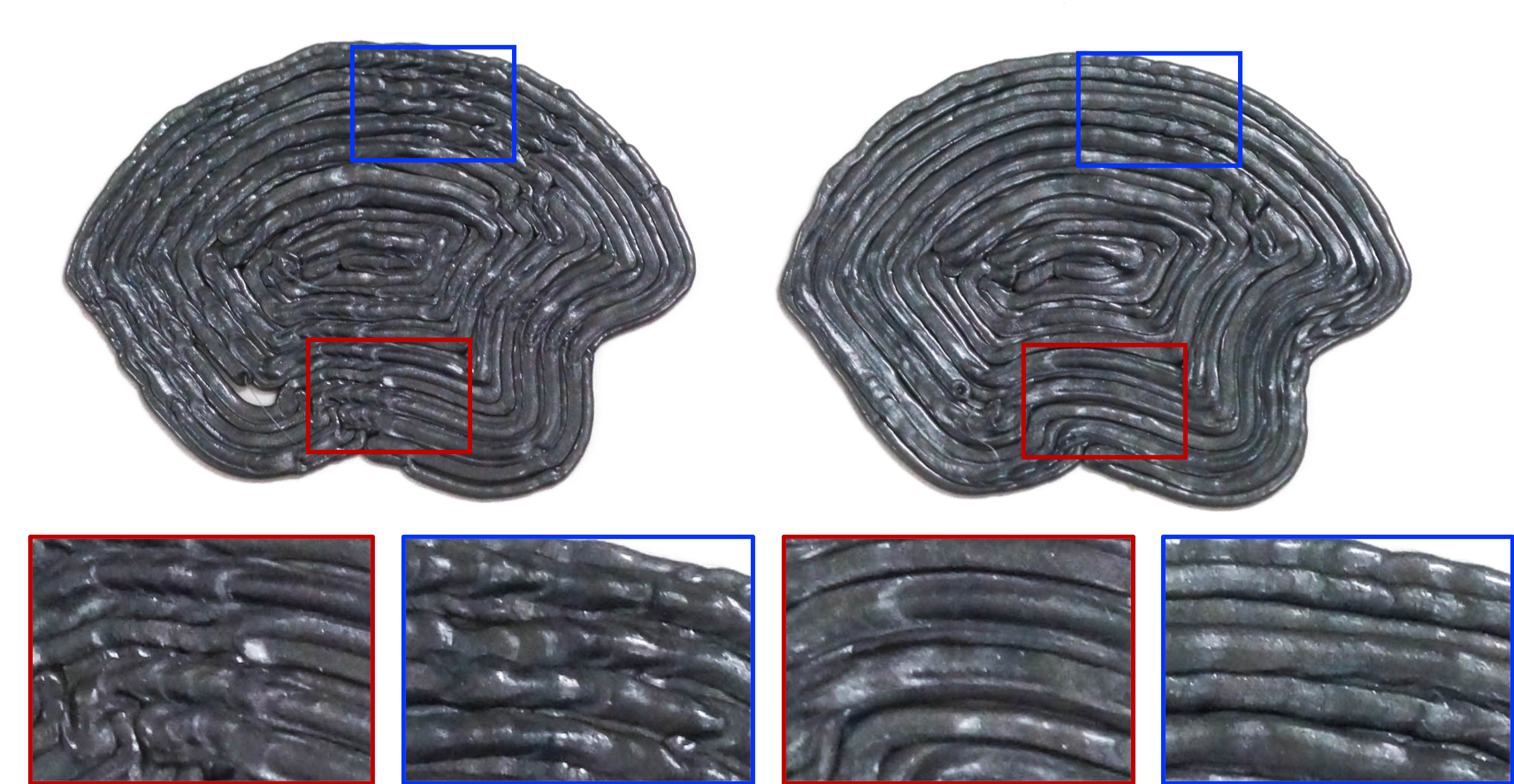}
\caption{The experimental test conducted on a robotic system with 8-DOFs (Fig.~\ref{figABBHardware}). (Top) The progressive results of jerk-optimization can effectively reduce the maximal jerk on all the eight joints to be less than a threshold $1.0$. (Bottom) The results of fabrication by a path with large jerk from graph-search (left) and a jerk-optimized path (right). Unwanted blobs can be observed on the result generated by a path with large jerk as the material deposition is not smooth. 
}\label{figFabricationResultA}
\end{figure}

We also test the trajectories computed by our method in physical experiments using robot-assisted 3D printing. 
To explicitly show the quality improvement in the real fabrication process, we choose an example tool-path for additive manufacturing on a planar layer as specimen. 
The specimen are fabricated on the \revision{8-DOFs}{8-DOF} system (Fig.~\ref{figABBHardware}), where the change of the maximum jerk at each joint during the iteration is shown at the top row of Fig.~\ref{figFabricationResultA}. 
The bottom row of Fig.~\ref{figFabricationResultA} gives the results of 3D printing by a graph-search based path (left) and the jerk-optimized path (right). Unwanted blobs can be clearly observed on the path with large jerks, while the jerk-optimized path leads to much smoother material deposition. For the sake of a better illustration, planar tool-paths are conducted in this experiment to demonstrate the influence of large jerk in additive manufacturing. 
The dynamic difference between the trajectories before and after optimization can be more clearly observed in the supplementary video at: \url{https://youtu.be/e8ISmh9MPrE}. 
In summary, the improvement of both the motion smoothness and the quality of fabrication that can be generated by our jerk-optimized trajectory planning algorithm is very significant. 

\revision{In terms of computational efficiency, the}{The} total time \revision{of}{required by} our trajectory planning approach is much shorter than the total time of 3D printing, which is a significant improvement compared to earlier work presented in \cite{dai18}. On average, a $20 \times$ speedup is achieved. For instance, the armadillo model shown in Fig.~\ref{figPath} contains more than 300 curved layers and more than 50 layers totalling $2000+$ waypoints. The original algorithm presented in \cite{dai18} needs about 40 hours for motion planning. With the help of the new algorithm presented in this paper, the total time for planning has been reduced to only 2 hours.
For a tool-path with $2,545$ waypoints, the computation can be completed in $116.43$ sec., which is much shorter than the fabrication time for the curved layer of this path -- i.e., around $460$ sec. 
Motion planning is no longer a bottleneck for robot-assisted fabrication as we can compute the trajectory of the next layer while working on the current one.

\subsection{\revision{}{Limitations}}
\begin{figure}[!t]
\includegraphics[width=\linewidth]{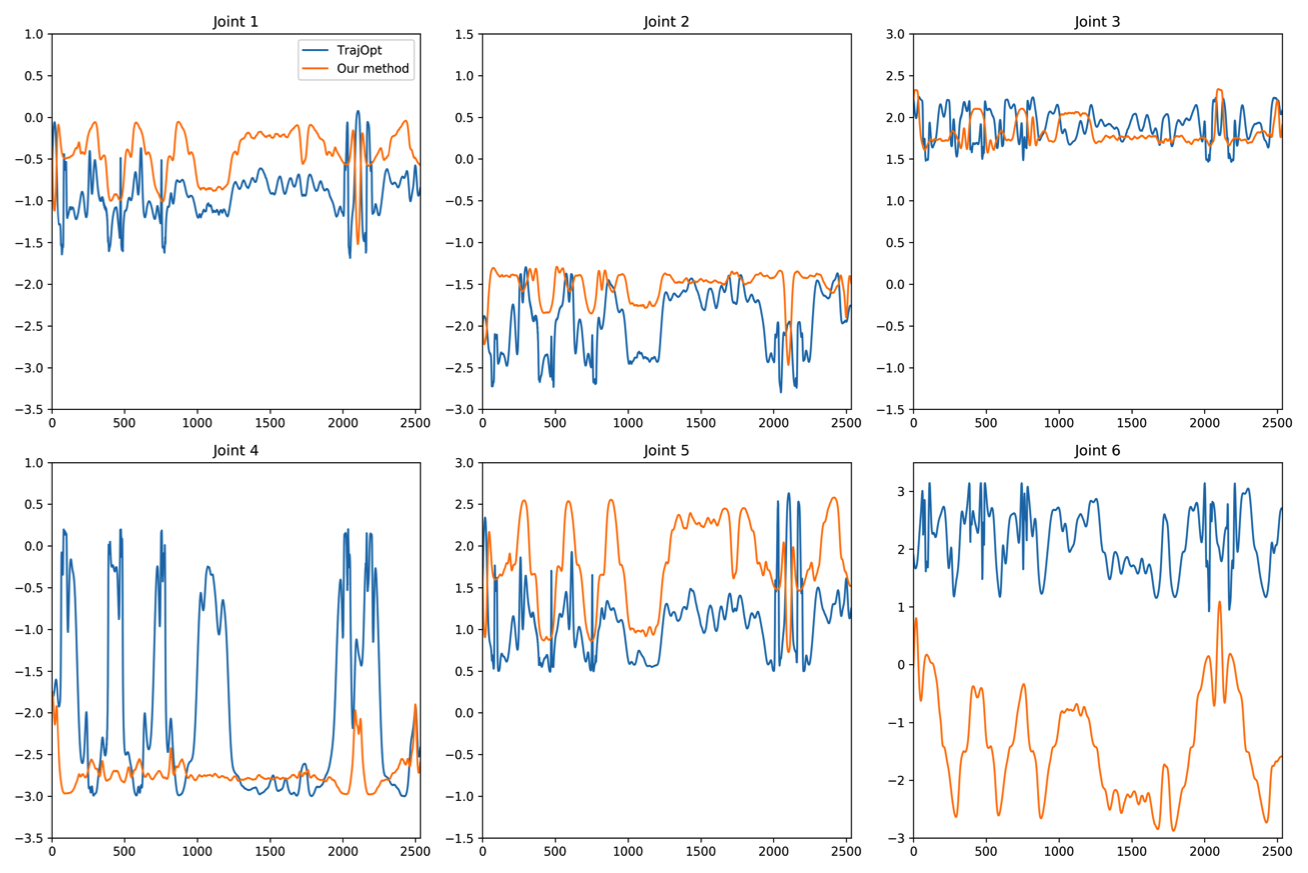}
\caption{\revision{}{When applying the TrajOpt approach~\cite{TrajOpt13} to the whole toolpath WS1, their best result is much worse than ours -- see the joint angles shown here.}
}\label{figCompareTrajOpt}
\end{figure}

\revision{}{Our method is an approach based on local processing so that a more optimal solution can be found by global methods (e.g., the TrajOpt approach~\cite{TrajOpt13})). When working on a toolpath with 60 waypoints (the first 60 points of WS1), the resultant trajectory with smaller total sum of squared jerks ($\mathbb{J}=1.64$) can be obtained from this TrajOpt method optimizing the whole path together. The result of our method is $\mathbb{J}=7.27$.
However, the major merit of our approach is its capability to handle a path with large number of waypoints, which is hard to be processed by existing methods. When applying the TrajOpt approach to a toolpath with many waypoints (e.g., the whole WS1), the optimizer is terminated by reaching the penalty iteration limit -- it means that the optimization is in fact unsuccessful. The best motion generated by TrajOpt is not well optimized (see Fig.\ref{figCompareTrajOpt} for a comparison with our approach).}

\begin{figure}[!t]
\includegraphics[width=\linewidth]{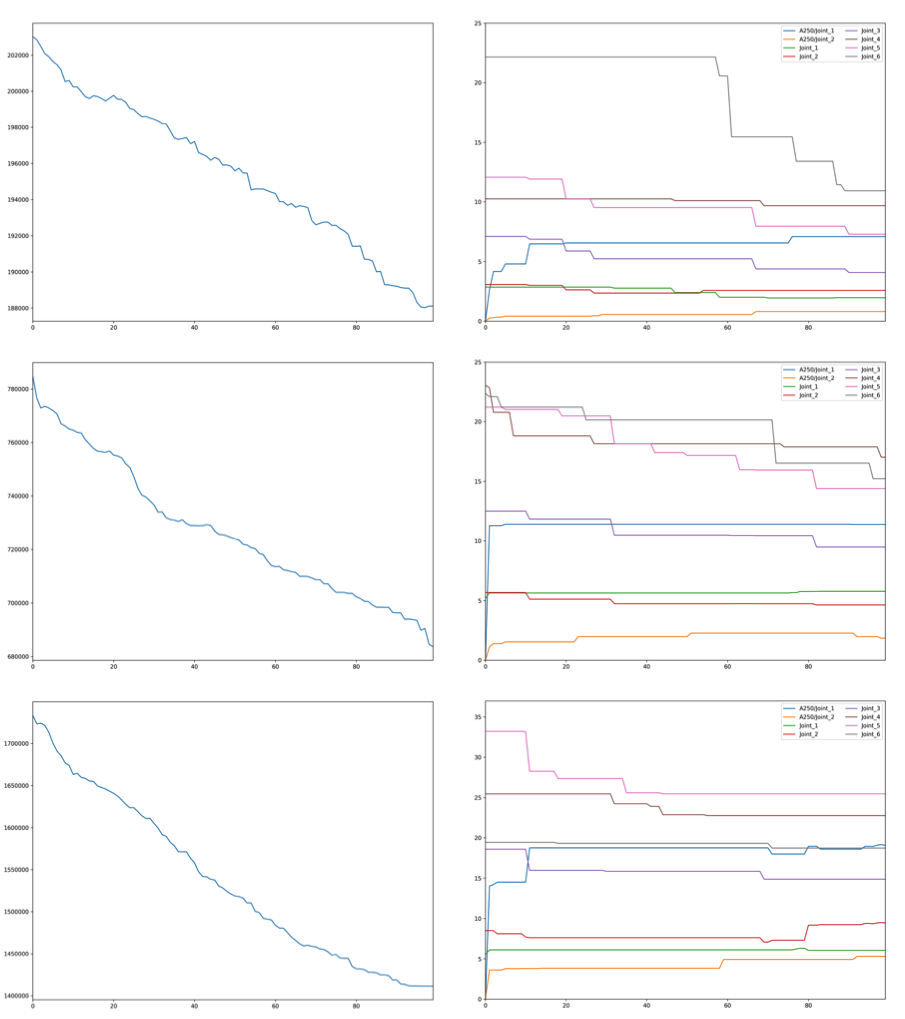}
\caption{\revision{}{Robustness tests conducted on the 8-DOF setup. Left column shows the total sum of squared jerks, $\mathbb{J}$ in Eq.(\ref{eqGlobalOptm}) and right shows the change of the maximum jerk at each joint during the iterations of our method, from top to bottom are with different noise level within 5, 10 and 15 degrees.}}\label{figNoise}
\end{figure}
\revision{}{It is also interesting to study the robustness of our approach by adding noises to the orientations of waypoints. Specifically, each orientation can be mapped to a point on the Gaussian sphere, and random noises are added within a range of 5, 10 and 15 degrees respectively in three tests taken on the 8-DOF system. The performance of our approach on noisy input is given in Fig.~\ref{figNoise}. The maximum jerk cannot meet the constraint as less than $1.0$ although the overall jerk $\mathbb{J}$ can always be reduced significantly. This is considered as the major limitation of our approach.}
%
%